\documentclass{article} 
\usepackage[final]{colm2025_conference}

\usepackage{tabularx}

\usepackage{setspace}

\usepackage[T1]{fontenc}

\usepackage{microtype}
\usepackage{hyperref}
\usepackage{url}
\usepackage{booktabs}
\usepackage{CJKutf8}
\usepackage{lineno}
\usepackage{graphicx}
\usepackage{multirow} 
\usepackage{float}
\usepackage{caption}
\usepackage{subcaption}
\usepackage{longtable}
\usepackage{subcaption}
\usepackage{booktabs}

\definecolor{darkblue}{rgb}{0, 0, 0.5}

\definecolor{darkgreen}{RGB}{0,170,0}

\hypersetup{colorlinks=true, citecolor=darkblue, linkcolor=darkblue, urlcolor=darkblue}

\title{Measuring Hong Kong Massive Multi-Task \\Language Understanding}

\author{Chuxue Cao$^{1*}$, Zhenghao Zhu$^{1*}$, Junqi Zhu$^{1}$, Guoying Lu$^{1}$, Siyu Peng$^{1}$,\\
\textbf{Juntao Dai}$^{2}$, \textbf{Weijie Shi}$^{1}$, \textbf{Sirui Han}$^{1\dag}$, \textbf{Yike Guo}$^{1\dag}$  \\
$^{1}$Hong Kong University of Science and Technology, 
$^{2}$Peking University,\\
\texttt{ccaoai@connect.ust.hk}
}

\begin{document}

\ifcolmsubmission
\linenumbers
\fi

\maketitle

\renewcommand{\thefootnote}{\fnsymbol{footnote}}
\footnotetext[1]{Equal contribution.
$^{\dag}$Corresponding authors.}

\begin{abstract}

\renewcommand{\thefootnote}{\arabic{footnote}} 

Multilingual understanding is crucial for the cross-cultural applicability of Large Language Models (LLMs). However, evaluation benchmarks designed for Hong Kong's unique linguistic landscape, which combines Traditional Chinese script with Cantonese as the spoken form and its cultural context, remain underdeveloped. To address this gap, we introduce HKMMLU, a multi-task language understanding benchmark that evaluates Hong Kong's linguistic competence and socio-cultural knowledge. The HKMMLU includes 26,698 multi-choice questions across 66 subjects, organized into four categories: Science, Technology, Engineering, and Mathematics (STEM), Social Sciences, Humanities, and Other. To evaluate the multilingual understanding ability of LLMs, 90,550 Mandarin-Cantonese translation tasks were additionally included. We conduct comprehensive experiments on GPT-4o, Claude 3.7 Sonnet, and 18 open-source LLMs of varying sizes on HKMMLU. The results show that the best-performing model, DeepSeek-V3, struggles to achieve an accuracy of 75\%, significantly lower than that of MMLU and CMMLU. This performance gap highlights the need to improve LLMs' capabilities in Hong Kong-specific language and knowledge domains. Furthermore, we investigate how question language, model size, prompting strategies, and question and reasoning token lengths affect model performance. We anticipate that HKMMLU will significantly advance the development of LLMs in multilingual and cross-cultural contexts, thereby enabling broader and more impactful applications. \footnote{The data are available at \url{https://huggingface.co/datasets/chuxuecao/HKMMLU}}

\end{abstract}

\section{Introduction}

Large Language Models (LLMs) such as GPT-4o, Claude 3.7 Sonnet, and Qwen-2.5 have garnered significant attention across various industries for their remarkable capability in various disciplines~\citep{openai2024gpt4ocard, anthropic2025claude37, qwen2025qwen25technicalreport}. Benchmarks have been developed to evaluate their performance and analyze their limitations~\citep{hendrycks2021, huang2023, chen2024measuring}. Massive Multitask Language Understanding (MMLU)~\citep{hendrycks2021} is a widely used English benchmark that evaluates LLMs across various subjects through multi-choice questions. Subsequently, similar studies have tried to extend MMLU to additional languages and regions, including Chinese~\citep{li2024cmmlu, tmmluplus_ikala2024improved, chen2024measuring, hsu2023}, Korean~\citep{son2024kmmlumeasuringmassivemultitask}, Indonesian and Spanish \citep{wang-etal-2024-seaeval}.

While benchmarks have been developed to evaluate models in the context of Chinese across various subjects, they are present in Simplified Chinese or Traditional Chinese~(Taiwan)~\citep{li2024cmmlu, huang2023, hsu2023, tmmluplus_ikala2024improved, chen2024measuring}. Consequently, challenges exist in evaluating knowledge and language specific to Hong Kong: 
(1) Hong Kong's socio-cultural uniqueness stems from its ``One Country, Two Systems'' policy and multicultural identity, blending local traditions and Western influences. However, CMMLU and C-Eval~\citep{li2024cmmlu, huang2023} focus on generalized Chinese contexts, neglecting Hong Kong’s distinct legal systems, historical narratives, and socio-linguistic practices. Consequently, these frameworks fail to evaluate LLMs' ability to understand region-specific knowledge and reasoning in multicultural scenarios; 
(2) Hong Kong Cantonese, as a spoken language, diverges markedly from the Mandarin-based written standard in Traditional Chinese script, preserving archaic grammar and regional vocabulary~\citep{cheng2014languagehood}, and cannot be directly translated word-for-word from Mandarin Chinese text. While TMMLU+ and TMLU~\citep{tmmluplus_ikala2024improved, chen2024measuring} evaluate the Traditional Chinese understanding capabilities of LLMs, the proficiency of language models in capturing Cantonese literary expression remains uncertain.

\begin{figure}[t]
    \centering
    \includegraphics[width=1\linewidth]{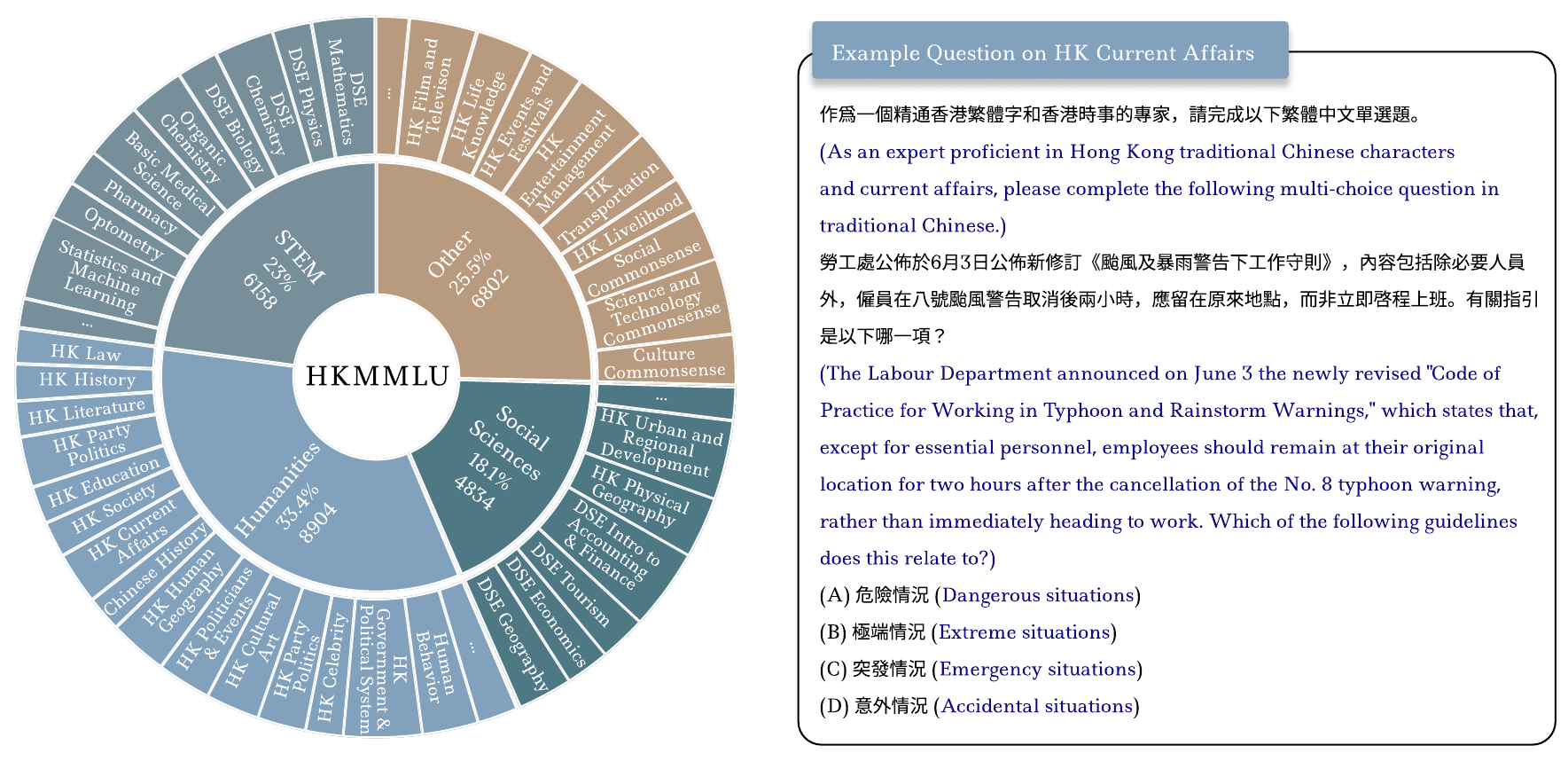}
    \caption{Overview of HKMMLU subject (partial) with an illustrative task example and bilingual English translations for clarity.}
    \label{fig:architecture}
\end{figure}
\setlength{\textfloatsep}{13pt}

To address these challenges, we introduce the Hong Kong Massive Multi-Task Language Understanding (HKMMLU) benchmark (Figure~\ref{fig:architecture}), specifically designed to evaluate LLMs in the unique linguistic and cultural context of Hong Kong. The benchmark consists of two parts: (1) a comprehensive knowledge base for Hong Kong consisting of 26,698 questions, divided into four categories: Science, Technology, Engineering, and Mathematics (STEM), Humanities, Social Sciences, and Other; (2) translation tasks for Cantonese, including 90,550 instances, with half of the translations from Mandarin to Cantonese and the other half in the opposite direction.

We evaluated GPT-4o, Claude 3.7 Sonnet, and 18 open-source LLMs across various model sizes using the HKMMLU benchmark. The results indicate that all models lack adequate knowledge about Hong Kong and Traditional Chinese, especially Cantonese, failing to achieve an accuracy of 75\%. For a deeper insight into the proficiency of the models in Cantonese literacy expression, we conducted experiments on translation tasks. The experimental results show that models performed better in Cantonese-to-Mandarin than in Mandarin-to-Cantonese, with Llama-3-Taiwan-70B-Instruct having the smallest gap. These findings underscore the importance of ongoing enhancement of Traditional Chinese and Hong Kong knowledge understanding abilities.

Furthermore, our extensive experiments reveal that:
(1) After translating HKMMLU to Simplified Chinese, the models exhibit a slight decrease in accuracy, with the average accuracy of all models dropping from 57\% to 56.7\%.
(2) Chain-of-thought (CoT) prompting improves model performance in STEM. Providing reasoning examples in the prompts further enhances model performance significantly in STEM and slightly in Social Sciences. 
(3) Few-shot prompting does not always enhance model performance. 
(4) Models with larger parameters within the same series tend to achieve higher accuracy. 
(5) When the question token length exceeds 600, the accuracy of most open-source models begins to decline, whereas the accuracy of closed-source models tends to improve. Additionally, the length of reasoning tokens is negatively correlated with model performance.
(6) We also conducted a human testing experiment involving 100 Hong Kong-specific questions. Even the top-performing model, DeepSeek-V3, failed to surpass human test-takers with a post-secondary degree, particularly on questions written in Cantonese.
\vspace{-2mm}

\section{Related Work}

Benchmarks are crucial for evaluating LLM capabilities, with many emerging to assess performance across various skills and subjects. For example, MATH~\citep{hendrycks2021measuring} and GSM8K~\citep{cobbe2021training} evaluated the mathematical ability of language models. Benchmarks like AI2~\citep{clark2018think}, CommonsenseQA~\citep{talmor2018commonsenseqa}, and Winogrande~\citep{sakaguchi2021winogrande} introduced the evaluation of commonsense reasoning. There are also benchmarks specifically designed to evaluate the reading comprehension~\citep{rajpurkar2018know, kwiatkowski2019natural, li2022multispanqa} and code generation capabilities of LLMs~\citep{chen2021evaluating, austin2021programsynthesislargelanguage}.

Besides evaluating the basic skills of LLMs, researchers have focused on different languages. In the early stages, benchmarks like GLUE~\citep{wang-etal-2018-glue} and SuperGLUE~\citep{wang2019superglue} have emerged for evaluating English natural language understanding. 
Additionally, numerous evaluation benchmarks for Chinese languages have been proposed, including SuperCLUE~\citep{xu2023supercluecomprehensivechineselarge} for natural language understanding, CMATH~\citep{wei2023cmath} for elementary school math, and MMCU~\citep{zeng2023measuringmassivemultitaskchinese} for medicine and education.
Notably, ACLUE~\citep{zhang-li-2023-large} evaluates ancient Chinese language ability, and AGIEVAL~\citep{zhong-etal-2024-agieval} offers evaluation in both Chinese and English for various standardized competitions and exams. In addition, several Traditional Chinese Benchmarks, such as DRCD~\citep{shao2019drcdchinesemachinereading} and TTQA~\citep{ennen2023extendingpretrainingbloomimproved}, assess reading comprehension and local commonsense knowledge in Taiwan.

Large-scale multitask evaluation benchmarks were introduced to provide a more comprehensive evaluation of LLMs. One prominent example is MMLU~\citep{hendrycks2021}, an English-only, multi-domain evaluation benchmark. After that, Simplified-Chinese benchmarks, such as CMMLU~\citep{li2024cmmlu} and C-EVAL~\citep{huang2023}, were developed for Chinese-specific subjects, while M3KE~\citep{liu2023m3kemassivemultilevelmultisubject} mainly focuses on the Chinese education sector. In addition, TMMLU~\citep{hsu2023} and TMMLU+~\citep{tmmluplus_ikala2024improved} evaluate model performance in a Traditional Chinese context. 
Furthermore, benchmarks like KMMLU~\citep{son2024kmmlumeasuringmassivemultitask} address the Korean language and local knowledge, while Cross-MMLU~\citep{wang-etal-2024-seaeval} integrates multiple languages and cultural perspectives. 

While these Chinese benchmarks generally evaluate models across various subjects in human society, they primarily focus on Simplified Chinese or Traditional Chinese (Taiwan), which differs from the language system used in Hong Kong, particularly Cantonese. Although HKCanto-Eval~\citep{cheng2025hkcantoevalbenchmarkevaluatingcantonese} has developed a Cantonese Benchmark, 78.4\% of its multi-choice tasks are translated from MMLU, with few tasks focusing on Hong Kong socio-culture. In contrast, most HKMMLU data is sourced from materials specific to Hong Kong, covering 23 subjects (14,912 data points), including topics such as Hong Kong Party Politics and Hong Kong Law.

\vspace{-1mm}
\section{HKMMLU}

\textbf{Task Overview} We introduce HKMMLU, a benchmark for evaluating the Traditional Chinese understanding and reasoning capabilities of LLMs. The benchmark covers diverse areas of knowledge, including the STEM, Social Sciences, Humanities, and Other domains. Except for non-region-specific subjects like mathematics, biology, and physics, the HKMMLU includes a wide range of Hong Kong-specific questions, such as physical geography, history, and law of Hong Kong. These questions evaluate Hong Kong-related knowledge of LLMs. Additionally, it features a Cantonese-Mandarin translation task.

\textbf{Data Collection} We collected multi-choice questions from primary and secondary school exams, Hong Kong Diploma of Secondary Education Examination (HKDSE), Hong Kong Knowledge Challenge questions, civic quiz competition questions, and so on, all of which have undergone rigorous manual annotation and selection. We also included some non-Taiwan-related questions from the TMMLU+, which constitutes 30.6\% of HKMMLU. Taiwan Traditional Chinese characters are converted to Hong Kong Traditional Chinese characters using \textit{OpenCC}~\citep{openccPython}. Additionally, we created the translation task manually.

\textbf{Data Processing} Each question is presented in a multi-choice format with two to three answer options. An example of our multi-choice question is shown in Figure~\ref{fig:architecture}. For questions without a subject during collection, we used three LLMs for labeling, including DeepSeek-V3~\citep{deepseekai2025deepseekv3technicalreport}, GPT-4o~\citep{openai2024gpt4ocard} and Qwen-2.5-72B-Instruct~\citep{qwen2025qwen25technicalreport}, and applied a majority voting approach. For each subject, one data sample was annotated using o3-mini~\citep{o3mini2025openai} to generate a reasoning process.

\textbf{Quality Control} To ensure the accuracy of labeling, we have manually checked 200 labels, and the accuracy rate is 96\%. For multi-choice questions that are converted from question-answer pairs, we have leveraged three processing models, including GPT-4o, DeepSeek-V3, and Claude 3.7 Sonnet, with a processing ratio of 1:1:1 for fairness. To secure the quality, we have manually checked 100 questions processed by each LLM; the accuracy rate is 97.7\%. 

\begin{figure}[htbp]
    \begin{center}
        \begin{minipage}{0.48\linewidth}
            \centering
            \includegraphics[width=\linewidth]{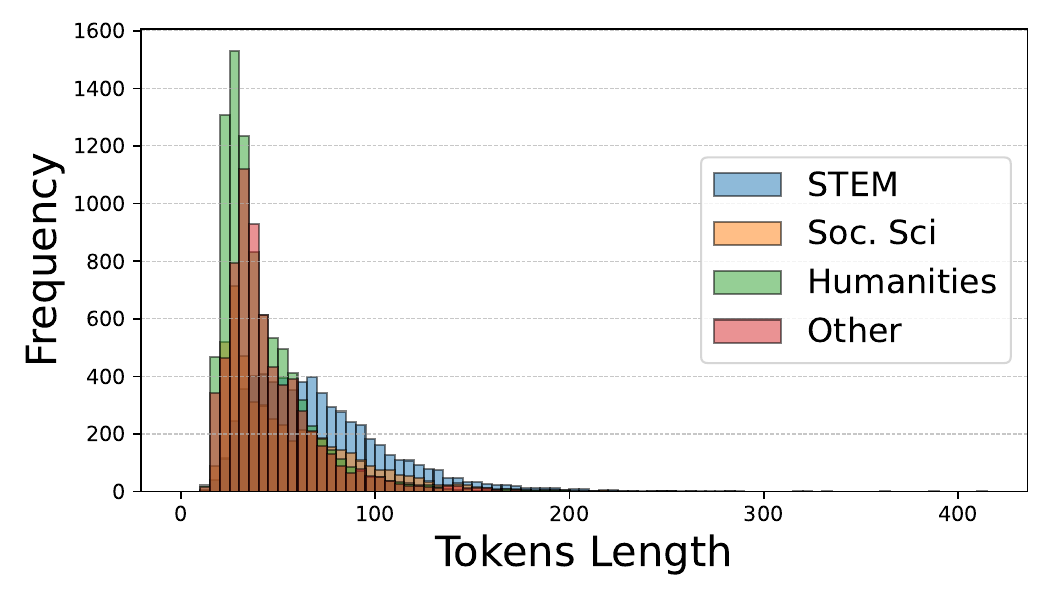} 
            \caption{Question~(include choices)~token length distribution of multi-choice questions in HKMMLU.}
            \label{fig:question-token-length}
        \end{minipage}%
        \hfill
        \begin{minipage}{0.48\linewidth}
            \centering
            \includegraphics[width=\linewidth]{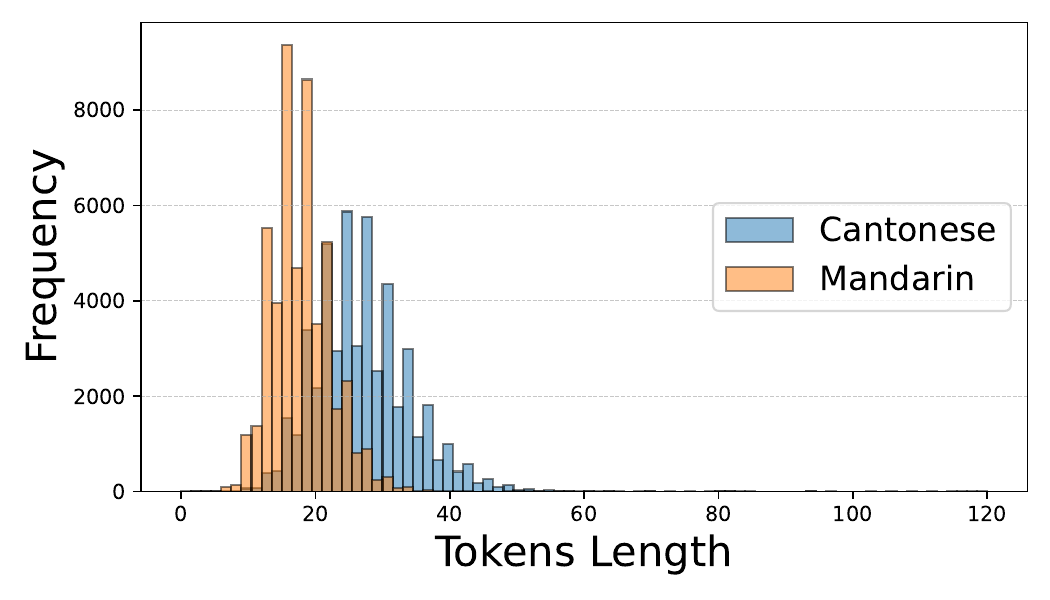} 
            \caption{Token length distribution of Cantonese and Mandarin sentences of translation tasks in HKMMLU.}
            \label{fig:translation-token-length}
        \end{minipage}
    \end{center}
\end{figure}

\textbf{Data Statistics} The HKMMLU dataset consists of four main categories: STEM, Humanities, Social Sciences, and Other, with a total of 66 sub-tasks and 26,698 data points. Figure~\ref{fig:question-token-length} illustrates that the token length of most multi-choice questions ranges from 30 to 200, with the STEM category featuring relatively longer questions. Figure~\ref{fig:translation-token-length} displays the distribution of sentence token lengths, indicating that Cantonese expressions tend to have slightly longer tokens due to the characteristics of the language. Refer to Table~\ref{tab:statistics} and Table~\ref{tab:subject_details} for detailed statistical results and subject information for HKMMLU.

\vspace{-3mm}

\section{Experiments}

\textbf{Setup} For multi-choice tasks, we evaluate the model in the following methods: 1) Direct Answer Evaluation: Models are prompted with questions directly; 2) CoT Evaluation: Models are asked to provide a reasoning process before giving their final answer; 3) Few-shot Evaluation: Models are provided with several question answering examples before the question; 4) 1-shot CoT Evaluation: A reasoning process for the example question is presented prior to the question. We use regular expressions to extract the answers and calculate the percentage of correct answers. 
For translation tasks, we directly prompt the LLM to translate the sentences and utilize BLEU~\citep{papineni-etal-2002-bleu}, METEOR~\citep{lavie-agarwal-2007-meteor}, and ROUGE-L~\citep{lin-2004-rouge} for evaluation.

\textbf{Models} We evaluated 20 models selected from 10 model families, including various model sizes. For closed-source models, we assessed GPT-4o~\citep{openai2024gpt4ocard} and Claude 3.7 Sonnet~\citep{anthropic2025claude37}. For the open-source models, we evaluated models including DeepSeek-V3~\citep{deepseekai2025deepseekv3technicalreport}, Gemma-2-it-2b/27b~\citep{gemmateam2024gemma2improvingopen}, GLM-4-9b-chat~\citep{glm2024chatglmfamilylargelanguage}, Llama-3.1-Instruct-8B/70B~\citep{llama31modelcard}, Llama-3-Taiwan-Instruct-8B/70B~\citep{lin2023taiwan}, Meta-Llama-3-Instruct-8B/70B~\citep{llama3modelcard}, Mistral-Instruct-Large-2411/Small-2409~\citep{jiang2023mistral7b}, Qwen2.5-Instruct-3B/7B/14B/72B~\citep{qwen2025qwen25technicalreport}, and Yi-1.5-Chat-9B/34B~\citep{young2024yi}.

\vspace{-1mm}
\subsection{Main Results}

\textbf{Results by model} The evaluation results of 0-shot prompting is shown in Table~\ref{tab:0shot_tc_sc_res}. We have observed the following: (1) DeepSeek-V3 achieves an average accuracy of 74.8\%, ranking first and surpassing both the two closed-source models. (2) Llama-3-Taiwan-Instruct series, fine-tuned on Traditional Chinese data, achieves the highest average accuracy among Llama models. Notably, the 70B variant of Llama-3-Taiwan-Instruct reaches an average accuracy of 67.4\%, outperforming Claude 3.7 Sonnet. (3) Among the remaining tested models, those fine-tuned on simplified Chinese data (e.g., models from the GLM, Qwen, and Yi families) generally outperform other model families of comparable sizes, such as the Llama series (excluding the Llama-3-Taiwan-Instruct-8B / 70B models), Gemma and Mistral models. The Qwen2.5-72B-Instruct model achieves an accuracy of 69.1\%, which is notably higher than the Mistral-Large-Instruct model, even though the latter has almost 123 billion parameters. (4) In contrast, most of the other models exhibit relatively low average accuracies, with many failing to exceed 60\%. Among all the tested models, the Llama-3.1-8B-Instruct from the Llama family shows the lowest performance, with an average accuracy of only 40\%. Additionally, in all series, larger parameter-sized LLMs tend to achieve higher average scores and perform better across all domains.

\begin{figure}[htbp]
    \centering
    \includegraphics[width=1\linewidth]{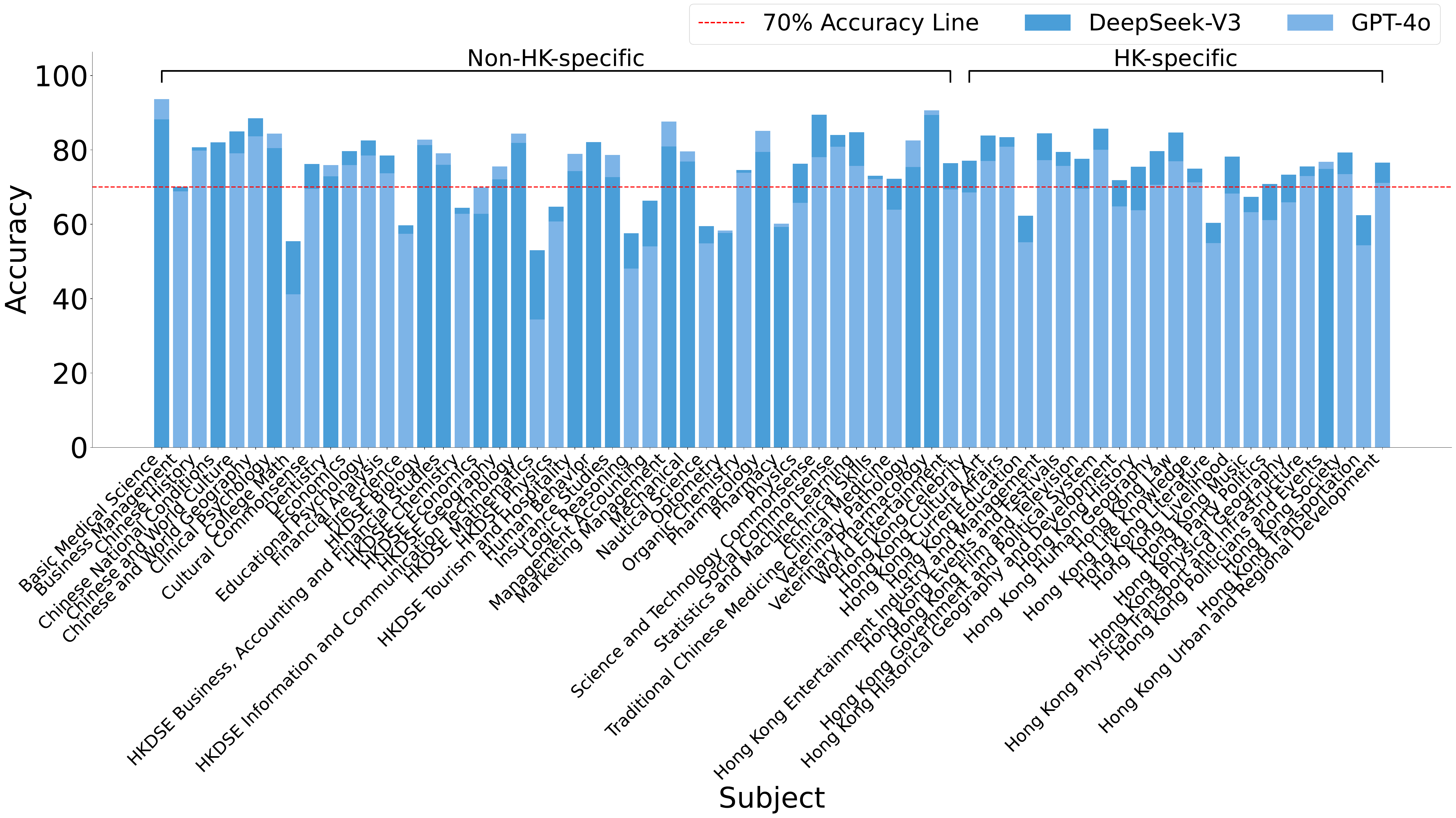}
    \caption{Comparing the two top-performing models.}
    \label{fig:comp-subject}
\end{figure}

\begin{table*}[!t]
    \centering
    \footnotesize
    \setlength{\tabcolsep}{4.5pt}
    \renewcommand{\arraystretch}{0.85}
      \resizebox{\columnwidth}{!}{

\begin{tabular}{l|cc|cc|cc|cc|cc}
    \toprule

    \multirow{2}{*}{\textbf{Model}} & \multicolumn{2}{c|}{\textbf{Avg.}} & \multicolumn{2}{c|}{\textbf{STEM}} & \multicolumn{2}{c|}{\textbf{Soc. Sci.}} & \multicolumn{2}{c|}{\textbf{Humanities}} & \multicolumn{2}{c}{\textbf{Other}} \\
    & tc~~&~sc & tc~~&~sc & tc~~&~sc & tc~~&~sc & tc~~&~sc  \\
    
    \midrule
    DeepSeek-V3 & \textbf{74.8} & \textbf{74.0} & \underline{74.5} & \underline{74.1} & \textbf{73.4} & \textbf{70.8} & \textbf{76.0} & \textbf{75.9} & \textbf{74.7} & \textbf{74.1} \\
    \midrule
    GPT-4o & \underline{70.3} & \underline{69.9} & \textbf{75.0} & \textbf{74.2} & \underline{70.4} & \underline{67.8} & 68.7 & \underline{70.1} & \underline{67.9} & \underline{67.1} \\
    \midrule
    Claude 3.7 Sonnet & 66.7 & 66.9 & 74.3 & 74.0 & 69.4 & 64.7 & 63.1 & 65.9 & 62.1 & 63.2 \\
    \midrule
    Gemma-2-2B-IT & 42.0 & 41.5 & 30.6 & 31.4 & 39.9 & 38.6 & 47.8 & 46.7 & 47.1 & 46.9 \\
    Gemma-2-27B-IT & 57.4 & 56.5 & 55.6 & 55.4 & 55.3 & 54.3 & 59.9 & 59.0 & 57.4 & 56.1 \\
    \midrule
    GLM-4-9B-Chat & 49.0 & 50.9 & 42.7 & 45.8 & 47.3 & 49.7 & 53.6 & 54.8 & 49.9 & 51.4 \\
    \midrule
    Llama-3.1-8B-Instruct & 40.0 & 41.8 & 32.6 & 34.4 & 38.1 & 38.1 & 44.5 & 47.4 & 42.7 & 44.5 \\
    Llama-3.1-70B-Instruct & 58.7 & 56.9 & 60.5 & 60.1 & 59.4 & 56.7 & 59.1 & 56.8 & 55.9 & 53.9 \\
    \midrule
    Llama-3-Taiwan-8B-Instruct & 51.9 & 48.4 & 47.2 & 45.9 & 48.9 & 45.4 & 56.7 & 52.1 & 52.4 & 48.2 \\
    Llama-3-Taiwan-70B-Instruct & 67.4 & 65.1 & 73.5 & 71.8 & 67.2 & 63.2 & 66.5 & 64.5 & 62.7 & 60.8 \\
    \midrule
    Llama-3-8B-Instruct & 45.1 & 44.7 & 33.3 & 35.6 & 41.2 & 41.0 & 53.3 & 51.0 & 48.7 & 48.1 \\
    Llama-3-70B-Instruct & 59.2 & 59.3 & 59.6 & 60.7 & 57.6 & 56.3 & 61.2 & 61.7 & 57.3 & 57.0 \\
    \midrule
    Mistral-Small-Instruct & 45.2 & 45.0 & 38.9 & 40.1 & 42.0 & 41.9 & 49.6 & 49.3 & 47.9 & 46.5 \\
    Mistral-Large-Instruct & 60.0 & 58.9 & 63.8 & 62.3 & 58.2 & 56.8 & 59.6 & 59.5 & 58.4 & 56.3 \\
    \midrule
    Qwen2.5-3B-Instruct & 50.3 & 51.4 & 42.5 & 46.0 & 47.8 & 49.4 & 54.4 & 53.5 & 54.7 & 55.4 \\
    Qwen2.5-7B-Instruct & 57.2 & 57.5 & 55.9 & 57.3 & 54.0 & 54.7 & 60.1 & 59.3 & 57.4 & 57.8 \\
    Qwen2.5-14B-Instruct & 62.3 & 64.1 & 63.8 & 65.2 & 61.5 & 61.4 & 62.9 & 65.7 & 60.5 & 63.1 \\
    Qwen2.5-72B-Instruct & 69.1 & 68.6 & 71.9 & 71.2 & 67.9 & 67.4 & \underline{70.2} & 69.8 & 65.9 & 65.5 \\
    \midrule
    Yi-1.5-9B-Chat & 52.6 & 52.7 & 45.5 & 48.8 & 51.6 & 50.2 & 57.0 & 55.4 & 54.3 & 55.0 \\
    Yi-1.5-34B-Chat & 60.6 & 59.7 & 56.2 & 55.4 & 58.7 & 56.3 & 64.7 & 64.5 & 60.7 & 60.3 \\
    \midrule
    Avg. & 57.0 & 56.7 & 54.9 & 55.5 & 55.5 & 54.2 & 59.4 & 59.1 & 56.9 & 56.6 \\
    
    \bottomrule
    \end{tabular}
    }
    
    \caption{Zero-shot performance of models on HKMMLU in Traditional Chinese (TC) and Simplified Chinese (SC). ``Soc. Sci'' stands for Social Sciences. ``Avg.'' indicates the micro-average accuracy. The highest score in each column is in bold, while the second highest score is underlined.} 
    \label{tab:0shot_tc_sc_res}
\end{table*}
\textbf{Result by subject} We compare the performance of the two best-performing models, including DeepSeek-V3 and GPT-4o, under 0-shot conditions. Results in Table~\ref{tab:0shot_tc_sc_res} reveal that except for a slightly lower score in the STEM category, DeepSeek-V3 significantly outperforms GPT-4o in Other categories. Notably, in the Humanities category, DeepSeek-V3 exceeds GPT-4o by 7.3 points. As detailed in Figure~\ref{fig:comp-subject}, DeepSeek-V3 demonstrates superior performance over GPT-4o in most subjects. Specifically, DeepSeek-V3 surpasses GPT-4o in Mathematics, Physics, and Chemistry, whereas GPT-4o shows higher accuracy in Biology, Medicine, and Pharmacy. This difference suggests that DeepSeek-V3 may have advantages in solving calculation-related questions. Furthermore, DeepSeek-V3 tends to perform better in subjects such as Hong Kong Current Affairs, Hong Kong Law, Hong Kong Politicians, and Events, which require a strong local knowledge base. In contrast, GPT-4o performs slightly better in subjects related to finance and economics.

\begin{table*}[h]
  \centering
  \fontsize{8.5pt}{8.5pt}\selectfont 
    \setlength{\tabcolsep}{3.4pt}
    \renewcommand{\arraystretch}{0.85}
    \resizebox{\columnwidth}{!}{
  \begin{tabular}{l|cc|cc|cc|cc}
    \toprule
    \textbf{Metric}  & \multicolumn{2}{c|}{\textbf{BLEU}} & \multicolumn{2}{c|}{\textbf{METEOR}} & \multicolumn{2}{c|}{\textbf{ROUGE-L}} & \multicolumn{2}{c}{\textbf{Avg.}}  \\
    \midrule
    \textbf{Task}  & C to M & M to C & C to M & M to C & C to M & M to C & C to M & M to C \\
    \midrule

\texttt{DeepSeek-V3}   & \textbf{57.4} & \underline{41.9} & \textbf{79.0} & \underline{67.1} & \textbf{80.2} & \underline{69.4} & \textbf{72.2} & \underline{59.5}  \\
\texttt{GPT-4o}        & \underline{55.9} & 20.9 & \underline{78.0} & 49.0 & \underline{79.3} & 53.5 & \underline{71.1} & 41.1  \\
\texttt{Claude 3.7 Sonnet} & 46.8 & 29.1 & 70.6 & 57.0 & 71.2 & 58.7 & 62.9 & 48.3 \\
\texttt{Gemma-2-2B-IT} & 31.4 & 8.7 & 61.2 & 29.1 & 66.4 & 37.9 & 53.0 & 25.2  \\
\texttt{Gemma-2-27B-IT}        & 46.0 & 20.5 & 72.6 & 47.9 & 75.2 & 53.7 & 64.6 & 40.7  \\
\texttt{GLM-4-9B-Chat} & 45.8 & 9.6 & 72.5 & 34.3 & 74.7 & 40.3 & 64.3 & 28.1  \\
\texttt{Llama-3.1-8B-Instruct} & 25.1 & 8.2 & 52.2 & 26.8 & 55.3 & 30.5 & 44.2 & 21.8  \\
\texttt{Llama-3.1-70B-Instruct}        & 38.4 & 13.4 & 65.1 & 36.8 & 67.4 & 42.7 & 57.0 & 31.0  \\
\texttt{Llama-3-Taiwan-8B-Instruct}    & 30.6 & 25.8 & 58.9 & 51.0 & 61.7 & 55.6 & 50.4 & 44.1  \\
\texttt{Llama-3-Taiwan-70B-Instruct}   & 48.2 & \textbf{45.1} & 72.6 & \textbf{69.8} & 74.0 & \textbf{71.4} & 64.9 & \textbf{62.1}  \\
\texttt{Llama-3-8B-Instruct}   & 42.4 & 19.2 & 68.8 & 43.2 & 70.9 & 49.1 & 60.7 & 37.2  \\
\texttt{Llama-3-70B-Instruct}  & 50.1 & 13.7 & 74.0 & 38.0 & 75.6 & 43.6 & 66.6 & 31.8  \\
\texttt{Mistral-Large-Instruct}        & 49.2 & 20.8 & 74.5 & 47.7 & 76.1 & 51.9 & 66.6 & 40.1  \\
\texttt{Mistral-Small-Instruct}        & 45.4 & 24.4 & 71.3 & 50.7 & 73.7 & 55.6 & 63.5 & 43.6  \\
\texttt{Qwen2.5-3B-Instruct}   & 38.6 & 5.3 & 65.5 & 24.8 & 68.3 & 31.3 & 57.5 & 20.5  \\
\texttt{Qwen2.5-7B-Instruct}   & 46.6 & 9.1 & 72.0 & 31.6 & 73.7 & 37.3 & 64.1 & 26.0  \\
\texttt{Qwen2.5-14B-Instruct}  & 49.3 & 21.1 & 73.7 & 47.4 & 74.9 & 51.6 & 66.0 & 40.0  \\
\texttt{Qwen2.5-72B-Instruct}  & 53.1 & 25.7 & 76.1 & 52.6 & 77.3 & 56.1 & 68.8 & 44.8  \\
\texttt{Yi-1.5-9B-Chat}        & 42.1 & 7.6 & 69.7 & 29.3 & 71.4 & 36.3 & 61.1 & 24.4  \\
\texttt{Yi-1.5-34B-Chat}       & 33.0 & 12.2 & 50.4 & 37.5 & 51.2 & 43.2 & 44.9 & 31.0  \\
\midrule

Avg.  & 43.6 & 18.6 & 68.8 & 42.9 & 70.9 & 47.9 & 61.1 & 36.5  \\

    \bottomrule
  \end{tabular}
  }
  \caption{\label{tab:translation} 
    Model performance on translation tasks. Models generally perform better in translating from Cantonese to Mandarin than from Mandarin to Cantonese. ``C'' stands for Cantonese, ``M'' stands for Mandarin. The highest score in each column is in bold, while the second highest score is underlined.
  }
\end{table*}
\textbf{Performance on Translation Tasks} We evaluated various models for translating between Mandarin and Cantonese, as shown in Table~\ref{tab:translation}. DeepSeek-V3 achieves the best performance in the Cantonese-to-Mandarin translation task across all evaluation metrics. However, it ranks second in the Mandarin-to-Cantonese translation task, with Llama-3-Taiwan-70B-Instruct ranking first. GPT-4o ranks second in the Cantonese-to-Mandarin translation task but performs poorly in the Mandarin-to-Cantonese task. This imbalanced performance, where models generally perform better in Cantonese-to-Mandarin translation than in Mandarin-to-Cantonese, is also observed in other models, with the Llama-3-Taiwan series exhibiting the smallest gap. This discrepancy may be attributed to the fact that Mandarin is more widely used in China and has more training materials available, leading to better performance in translating from Cantonese to Mandarin. These results emphasize the benefits of fine-tuning with traditional Chinese materials. However, it is noteworthy that the performance of fine-tuned models in the Cantonese-to-Mandarin translation task has decreased, suggesting that while fine-tuning can improve overall translation capabilities, it may not uniformly enhance performance across all translation directions.

\vspace{-2mm}
\subsection{Analysis}

\subsubsection{Performance on Traditional Chinese and Simplified Chinese}

To evaluate the performance of LLMs in Simplified Chinese, we translated the HKMMLU into Simplified Chinese using GPT-4o. Contrary to the conclusions drawn from TMMLU+~\citep{tmmluplus_ikala2024improved}, which suggest that non-Traditional Chinese LLMs perform better in Simplified Chinese, the results from HKMMLU indicate that most LLMs achieve higher average scores in Traditional Chinese. Specifically, most models excel in Traditional Chinese in Social Sciences, Humanities, and Other categories, while their performance in STEM is comparatively weaker. This discrepancy may stem from the fact that the first three categories include more Hong Kong-specific knowledge, which is more frequently represented in Traditional Chinese within the training materials of LLMs, reflecting the unique cultural context of Hong Kong. In contrast, STEM topics such as Mathematics and Biology are more general and often use standardized terminologies.

\begin{figure}[htbp]
    \begin{center}
        \begin{minipage}{0.48\linewidth}
            \centering
            \includegraphics[width=\linewidth]{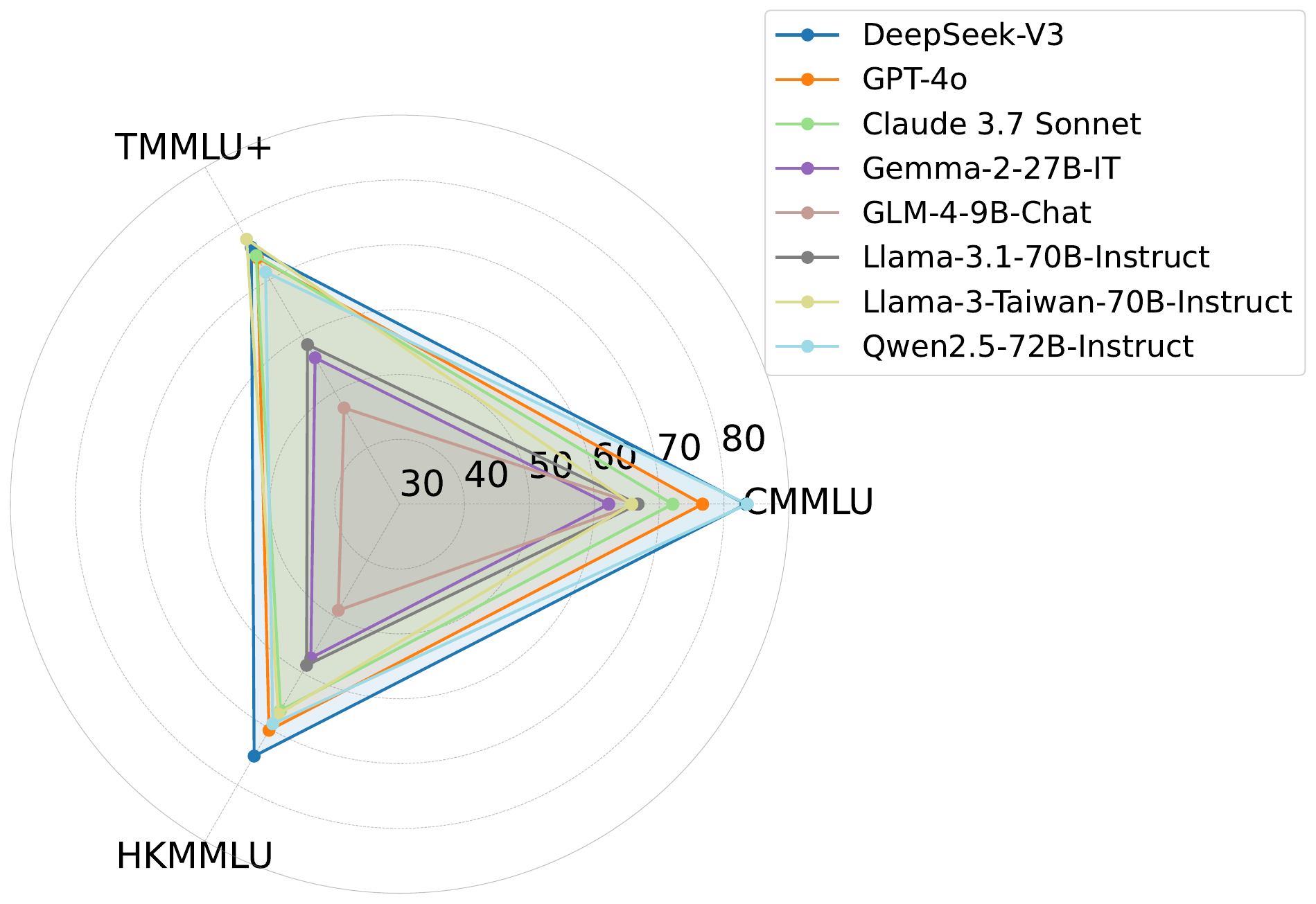} 
            \caption{Comparison of model performance among HKMMLU, TMMLU+, and CMMLU, with most models achieving their best results on CMMLU.}
            \label{fig:mmlu_compare}
        \end{minipage}
        \hfill
        \begin{minipage}{0.48\linewidth}
            \centering
            \includegraphics[width=\linewidth]{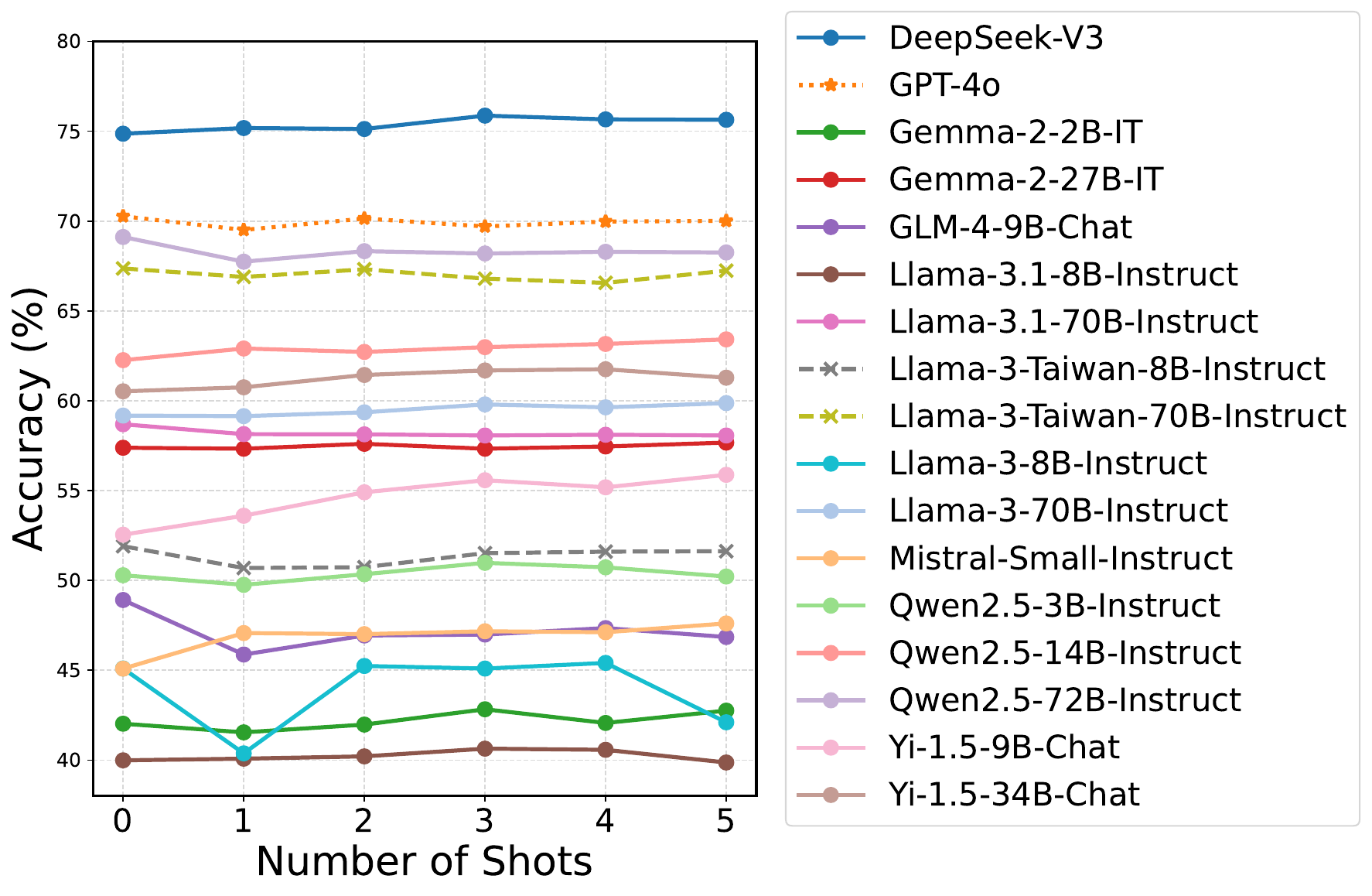} 
            \caption{A comparison of zero-shot and few-shot prompts on accuracy on HKMMLU.}
            \label{fig:few_shot}
        \end{minipage}
    \end{center}
\end{figure}

We also compare the performance of different models on different MMLU benchmarks. As shown in Figure~\ref{fig:mmlu_compare}, non-Traditional Chinese models, such as DeepSeek-V3 and Qwen2.5-72B-Instruct, perform worse on TMMLU+ and HKMMLU compared to CMMLU, indicating their deficiency on Traditional Chinese understanding. Llama-3-Taiwan-70B-Instruct, fine-tuned with Traditional Chinese materials related to Taiwan, performs best on TMMLU+. However, although Hong Kong and Taiwan Traditional Chinese share most characters, Llama-3-Taiwan-70B-Instruct performs poorly on HKMMLU, indicating a lack of knowledge of Hong Kong and Cantonese.

\subsubsection{Efficiency of Chain-of-Thought (CoT) Prompting}

To evaluate the effectiveness of CoT prompting, we first applied a 0-shot CoT prompt as detailed in Appendix~\ref{app:inference}, which requires the LLM to think step-by-step to generate an answer. Similar to the conclusions drawn from CMMLU~\citep{li2024cmmlu}, the average scores of all models drop significantly. However, models such as DeepSeek-V3, Gemma 2 series, Llama-3 series, and Yi-1.5 series show improved performance in the STEM category. This improvement is because the STEM category includes subjects such as mathematics and physics, which may require certain reasoning steps to generate the correct answer. 

We further investigate whether an example reasoning step would encourage the LLMs to generate more convincing answers. As shown in Table~\ref{tab:cot_res}, several have shown significant improvement despite the decreased average accuracy for most models, such as DeepSeek-V3 and GLM-4-9B-Chat. Notably, DeepSeek-V3, GPT-4o, and GLM-4-9B-Chat increased by more than five percentage points in the STEM category. The significant improvement in performance of these models in 1-shot CoT compared to 0-shot CoT can be attributed to the high-quality reasoning example provided. This finding further proves the necessity of reliable reasoning steps for tasks in STEM. These models also show slight enhancement in Social Sciences. However, only GPT-4o and Llama-3-70B-Instruct show an increase in scores in the Other subject, while no model shows improvement in Humanities. This limitation is because these two categories include more Hong Kong-specific knowledge, which cannot be learned through reasoning if the models lack relevant knowledge.

\begin{table*}[!t]
    \centering
    \small
    \setlength{\tabcolsep}{4.5pt}
    \renewcommand{\arraystretch}{0.9}
    \resizebox{\textwidth}{!}{
    \begin{tabular}{l|ccc|ccc|ccc|ccc|ccc}
    \toprule

     \multirow{2}{*}{\textbf{Model}} 
     & \multicolumn{3}{c|}{\textbf{Avg.}} & \multicolumn{3}{c}{\textbf{STEM}} & \multicolumn{3}{c}{\textbf{Soc. Sci.}} & \multicolumn{3}{c}{\textbf{Humanities}} & \multicolumn{3}{c}{\textbf{Other}} \\
    
    \footnotesize & DA~~&~0-CoT~~&~1-CoT & DA~~&~0-CoT~~&~1-CoT & DA~~&~0-CoT~~&~1-CoT & DA~~&~0-CoT~~&~1-CoT & DA~~&~0-CoT~~&~1-CoT  \\

    \midrule

    \texttt{DeepSeek-V3} & 74.9 & 74.1 & \underline{75.8} & 74.5 & 78.1 & \underline{81.0} & 73.5 & 74.3 & \underline{75.6} & \underline{76.0} & 72.9 & 74.3 & \underline{74.8} & 71.8 & 72.9\\
    \texttt{GPT-4o} & 70.3 & \underline{72.4} & 72.2 & 75.0 & 79.7 & \underline{82.1} & 70.4 & 74.1 & \underline{74.1} & 68.6 & \underline{69.1} & 66.6 & 67.9 & 68.3 & \underline{68.8} \\
    \texttt{Claude 3.7 Sonnet} & 66.7 & \underline{72.3} & 66.8 & 74.3 & \underline{81.7} & 78.6 & 69.4 & \underline{76.2} & 74.8 & 63.0 & \underline{67.3} & 58.7 & 62.1 & \underline{66.6} & 59.5 \\
    \texttt{Gemma-2-2B-IT} & \underline{42.0} & 36.6 & 36.8 & 30.6 & \underline{31.6} & 30.2 & \underline{40.1} & 35.9 & 34.4 & \underline{47.7} & 38.4 & 40.3 & \underline{47.0} & 39.8 & 40.4\\
    \texttt{Gemma-2-27B-IT} & \underline{57.4} & 55.8 & 56.9 & 55.7 & 59.2 & \underline{60.6} & 55.4 & 55.0 & \underline{57.7} & \underline{59.7} & 55.8 & 56.3 & \underline{57.5} & 53.3 & 53.4\\
    \texttt{GLM-4-9B-Chat} & 48.9 & 49.7 & \underline{51.0} & 42.8 & 51.0 & \underline{51.2} & 47.3 & 48.6 & \underline{50.6} & \underline{53.4} & 50.1 & 52.3 & \underline{49.9} & 49.0 & 49.3\\
    \texttt{Llama-3.1-8B-Instruct} & \underline{40.0} & 34.0 & 29.8 & 32.6 & \underline{32.9} & 29.0 & \underline{38.1} & 34.8 & 30.8 & \underline{44.4} & 34.9 & 30.2 & \underline{42.7} & 33.4 & 29.0\\
    \texttt{Llama-3.1-70B-Instruct} & \underline{58.7} & 49.8 & 39.6 & \underline{60.5} & 51.9 & 43.2 & \underline{59.5} & 50.7 & 41.4 & \underline{58.8} & 49.5 & 36.9 & \underline{55.9} & 47.4 & 38.3\\
    \texttt{Llama-3-Taiwan-8B-Instruct} & \underline{51.9} & 49.6 & 51.2 & 47.4 & 45.5 & \underline{48.3} & 48.9 & 47.7 & \underline{49.3} & \underline{56.6} & 53.1 & 53.8 & \underline{52.4} & 50.2 & 51.9\\
    \texttt{Llama-3-Taiwan-70B-Instruct} & \underline{67.4} & 64.9 & 66.5 & 73.6 & 72.8 & \underline{74.4} & \underline{67.3} & 65.5 & 67.2 & \underline{66.3} & 62.5 & 64.0 & \underline{62.6} & 60.0 & 61.5\\
    \texttt{Llama-3-8B-Instruct} & 45.1 & 40.6 & \underline{45.8} & 33.3 & 35.5 & \underline{37.6} & 41.3 & 39.4 & \underline{42.5} & \underline{53.2} & 43.6 & 52.1 & \underline{48.6} & 42.7 & 47.9\\
    \texttt{Llama-3-70B-Instruct} & 59.2 & 58.2 & \underline{60.5} & 59.7 & 60.3 & \underline{62.4} & 57.8 & 57.9 & \underline{61.3} & \underline{61.0} & 57.6 & 60.0 & 57.2 & 57.4 & \underline{58.6}\\
    \texttt{Mistral-Small-Instruct} & \underline{45.1} & 42.3 & 40.2 & 38.9 & \underline{43.0} & 40.2 & \underline{42.1} & 39.6 & 37.9 & \underline{49.4} & 43.0 & 41.9 & \underline{47.8} & 43.2 & 39.9\\
    \texttt{Mistral-Large-Instruct} & \underline{60.0} & 58.7 & 49.3 & \underline{63.8} & 59.2 & 51.7 & \underline{58.2} & 56.3 & 50.8 & 59.5 & \underline{59.9} & 48.1 & 58.4 & \underline{58.6} & 47.2\\
    \texttt{Qwen2.5-3B-Instruct} & \underline{50.3} & 42.1 & 44.0 & \underline{42.5} & 41.5 & 41.8 & \underline{48.0} & 40.5 & 42.4 & \underline{54.1} & 41.9 & 44.8 & \underline{54.7} & 44.5 & 46.6\\
    \texttt{Qwen2.5-7B-Instruct} & \underline{57.2} & 53.1 & 53.9 & 56.0 & 56.0 & \underline{56.3} & \underline{54.1} & 51.8 & 52.7 & \underline{60.0} & 52.6 & 54.0 & \underline{57.4} & 51.9 & 52.4\\
    \texttt{Qwen2.5-14B-Instruct} & \underline{62.3} & 61.2 & 61.7 & 63.9 & \underline{65.2} & 64.7 & 61.5 & 61.3 & \underline{61.8} & \underline{62.8} & 60.5 & 61.5 & \underline{60.4} & 58.1 & 58.8\\
    \texttt{Qwen2.5-72B-Instruct} & \underline{69.1} & 66.9 & 66.6 & \underline{72.0} & 71.4 & 71.1 & \underline{67.9} & 67.6 & 67.3 & \underline{70.1} & 65.8 & 65.6 & \underline{65.8} & 63.6 & 62.7\\
    \texttt{Yi-1.5-9B-Chat} & \underline{52.5} & 49.7 & 50.5 & 45.4 & \underline{45.9} & 45.4 & \underline{51.6} & 49.1 & 49.0 & \underline{56.9} & 51.3 & 54.2 & \underline{54.3} & 51.9 & 51.5\\
    \texttt{Yi-1.5-34B-Chat} & \underline{60.5} & 56.1 & 58.1 & \underline{56.2} & 54.4 & 55.5 & \underline{58.7} & 56.5 & 57.1 & \underline{64.6} & 56.7 & 60.1 & \underline{60.5} & 56.6 & 58.7 \\
    
    \midrule
    \texttt{Avg.} & \underline{57.0} & 54.4 & 53.9 & 54.9 & \underline{55.8} & 55.3 & \underline{55.6} & 54.1 & 53.9 & \underline{59.3} & 54.3 & 53.8 & \underline{56.9} & 53.4 & 52.5 \\
    \bottomrule
    \end{tabular}
    }
    \caption{Model performance on HKMMLU using direct answering (DA), 0-shot CoT (0-CoT), and 1-shot CoT prompting (1-CoT). ``Soc. Sci'' stands for \textit{Social Sciences}. ``Avg.'' represents the micro-average accuracy. The highest score within each model and category across the three methods is underlined.} 
    \label{tab:cot_res}
\end{table*}

\subsubsection{Efficiency of Few-shot Prompting}

Contrary to previous studies suggesting that an increase in the number of shots leads to improved performance~\citep{NEURIPS2020_1457c0d6}, our findings indicate that the performance of most models tends to stabilize (Figure~\ref{fig:few_shot}). However, some models, such as Llama-3-8B-Instruct and GLM-4-9B-Chat, experience a sharp decline between 0-shot and 1-shot. In contrast, models like Yi-1.5-9B-Chat and Mistral-Small-Instruct show a relatively significant improvement from 0-shot to 1-shot. DeepSeek-V3 shows a slight increase in performance from 2-shot to 3-shot, maintaining its rank one position. The results indicate that the number of shots affects model performance differently across models and does not always lead to improved performance.

\subsubsection{Impact of Question Token Length and Reasoning Token Length}

\begin{figure}[htbp]
    \centering
    \includegraphics[width=\linewidth]{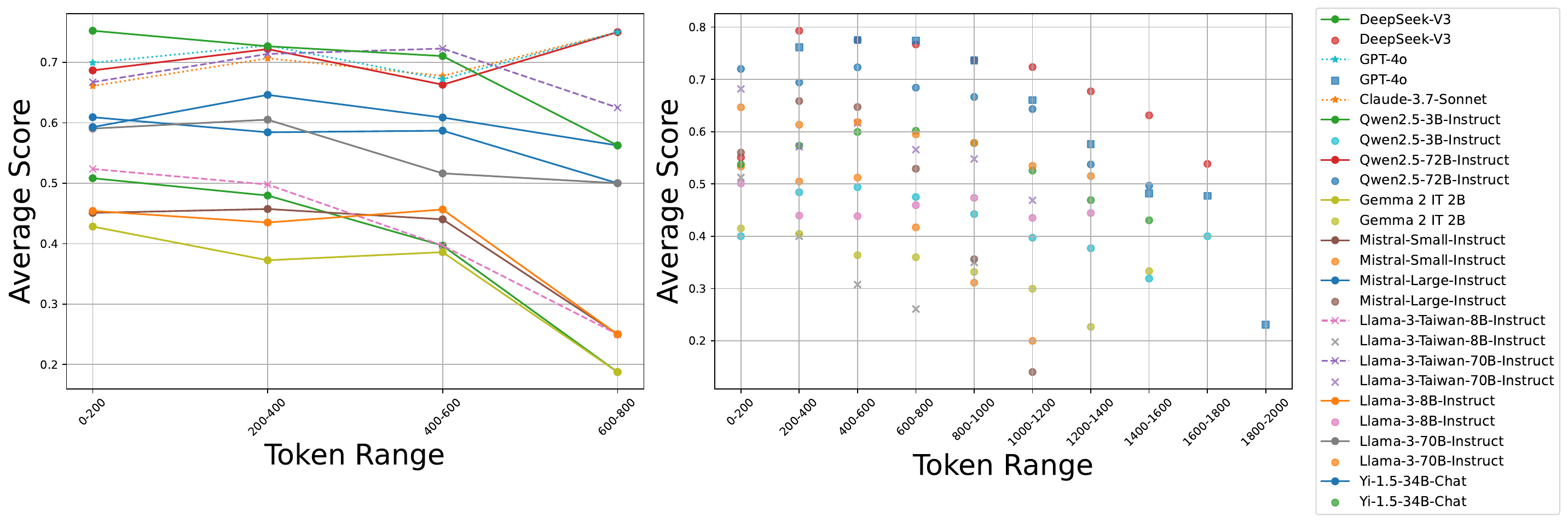}
    \caption{Impact of question token length on model performance (left) and impact of reasoning token length on model performance (right).}
    \label{fig:token-length}
\end{figure}

We explore the impact of the token length of questions on model performance. As shown in Figure~\ref{fig:token-length}, most models show a decrease in accuracy as the token length increases beyond 400, except the two closed-source models and Qwen-2.5-72B-Instruct. Even the top-performing model, DeepSeek-V3, shows a sharp decline after 600 token length. This trend highlights the deficiency of most open-source LLMs in understanding long texts in Traditional Chinese. Specifically, the two Traditional Chinese-fine-tuned models, Llama-3-Taiwan-70B-Instruct and Llama-3-Taiwan-8B-Instruct, do not show advantages in long text understanding. The 70B model declines after 600 tokens, while the 8B model declines after 400 tokens. To examine the relationship between the token length of reasoning and model performance, we record the accuracy of models at various reasoning token lengths when utilizing 1-shot CoT. Figure~\ref{fig:token-length} illustrates that models tend to underperform when the reasoning token length exceeds 400 tokens. This observation suggests that longer reasoning tokens do not necessarily enhance model accuracy without adequate knowledge.

\subsubsection{Comparative Analysis of Human and LLM Performance}

\begin{table*}[h]
    \centering
    \setlength{\tabcolsep}{4.5pt}
    \renewcommand{\arraystretch}{0.85}
    \resizebox{0.4\columnwidth}{!}{
    \begin{tabular}{l|c|cc}
    
    \toprule

Tester         & Avg.   & Cantonese      & TC \\
\midrule
DeepSeek-V3    & 57.0   & 59.2      & 56.2 \\
Human          & \textbf{57.9}   & \textbf{63.4}      & \textbf{55.9} \\

    \bottomrule
    \end{tabular}
    }
    \caption{Performance comparison between humans and the top LLM, DeepSeek-V3, on 100 selected Hong Kong-specific questions. ``TC'' denotes Traditional Chinese.} 
    \label{tab:human}
\end{table*}

To reduce over-reliance on LLM-driven analysis in the benchmark~\citep{yeadon2024comparison} and to assess their actual understanding of Traditional Chinese and Cantonese, we have implemented a human-machine comparison test.
We selected 100 questions from Hong Kong-specific subjects and tested them on 24 local testers with a post-secondary degree or above. Humans achieved an average accuracy of 57.9\% (SD = 5.9\%, SE = 1.2\%) with a 95\% confidence interval of [55.5\%, 60.2\%]. 
In comparison, the top-performing LLM, DeepSeek-V3, achieved an accuracy of 57\% (Table~\ref{tab:human}), slightly below the average accuracy of human test takers.
These results reveal the limitations of LLMs in matching the knowledge and language skills of local individuals while also highlighting the challenges of our benchmark. 
Furthermore, we also find that the gap between human and LLM is greater for questions in Cantonese than for those in Traditional Chinese, indicating DeepSeek-V3's deficiency in understanding Cantonese linguistic and cultural nuances.

\vspace{-2mm}
\section{Conclusion}

We introduce HKMMLU, a benchmark designed to evaluate the proficiency of LLMs in Hong Kong knowledge and its languages through multi-choice questions and Mandarin-Cantonese translation tasks. Experimental results highlight significant performance gaps in LLMs when addressing region-specific linguistic nuances and socio-cultural context. Our analysis further identifies key factors influencing model performance, including question language, model size, prompting strategies, and the lengths of questions and reasoning tokens. 
A comparison test between humans and the top-performing LLM, DeepSeek-V3, on 100 Hong Kong-specific questions further demonstrates the deficiencies of LLMs in knowledge of Hong Kong and its languages, particularly Cantonese.
Focusing on Hong Kong's unique sociolinguistic landscape, this work aims to drive advancements in LLMs' multilingual and cross-cultural capabilities, fostering AI systems that are both technically robust and culturally attuned to global and local needs.

\section*{Ethics Statement}
All data in HKMMLU are sourced from publicly available materials, ensuring transparency and accessibility. Humans have carefully reviewed each instance to ensure no harmful or offensive questions are included. Additionally, when applying the LLMs for labeling or inference, we did not use any damaging or intentionally provocative prompts that could lead to safety or unethical outputs.

\bibliography{colm2025_conference}
\bibliographystyle{colm2025_conference}

\clearpage
\appendix

\begin{spacing}{0.5}
\tableofcontents
\end{spacing}

\section{HKMMLU Subjects}
\renewcommand{\arraystretch}{1.1} 
\begin{table*}[htbp]
\scriptsize
\centering
\small

\begin{CJK}{UTF8}{bsmi}
\resizebox{1\textwidth}{!}{
\begin{tabular}{cllc}
\toprule
    \textbf{Category}  & \textbf{Subject Name} & \textbf{Traditional Chinese Name} & \textbf{Questions Number} \\ 
    
    \midrule
    
    STEM & 
    \begin{tabular}[c]{l}Basic Medical Science \\ Dentistry \\ College Math \\ HKDSE Biology \\ HKDSE Chemistry \\ HKDSE Information and Communication technology \\ HKDSE Mathematics \\ HKDSE Physics \\ Mechanical \\ Optometry \\ Organic Chemistry \\ Pharmacology \\ Pharmacy \\ Physics \\ Traditional Chinese Medicine Clinical Medicine \\ Veterinary Pathology \\ Veterinary Pharmacology \\ Statistics and Machine Learning \\ 
    \end{tabular} &
    \begin{tabular}[c]{l}基礎醫學 \\ 牙醫學 \\ 大學數學 \\ 香港中學文憑考試生物 \\ 香港中學文憑考試化學 \\ 香港中學文憑考試資訊與通訊科技 \\ 香港中學文憑考試數學 \\ 香港中學文憑考試物理 \\ 機械與機電概論 \\ 視光學 \\ 有機化學 \\ 藥理學 \\ 藥劑學 \\ 物理 \\ 中醫臨床醫學 \\ 獸醫病理學 \\ 獸醫藥理學 \\ 統計與機器學習 \\ 
    \end{tabular} 
    &
    \begin{tabular}[c]{c}1020 \\ 296 \\ 176 \\ 140 \\ 254 \\ 360 \\ 184 \\ 103 \\ 109 \\ 891 \\ 131 \\ 620 \\ 350 \\ 115 \\ 278 \\ 310 \\ 577 \\ 244 \\ 
    \end{tabular} \\ 
    
    \midrule
    
    Social Science & 
    \begin{tabular}[c]{l}Business Management \\ Chinese            and World Geography \\ Clinical Psychology \\ Economics \\ Educational Psychology \\ Financial Analysis \\ HKDSE Business, Accounting and Financial Studies \\ HKDSE Geography \\ HKDSE Tourism and Hospitality \\ HKDSE Economics \\ Hong Kong Physical Geography \# \\ Hong Kong Transport and Infrastructure \# \\ Hong Kong Urban and Regional Development \# \\ Insurance Studies \\ Management Accounting \\ Marketing Management \\ 
    \end{tabular} &
    \begin{tabular}[c]{l}企業管理 \\ 中國與世界地理 \\ 臨床心理學 \\ 經濟學 \\ 教育心理學 \\ 財務分析 \\ 香港中學文憑考試企業、會計與財務概論 \\ 香港中學文憑考試地理 \\ 香港中學文憑考試旅遊與款待 \\ 香港中學文憑考試經濟 \\ 香港自然地理 \\ 香港交通與基礎設施 \\ 香港城市與區域規劃 \\ 保險學 \\ 管理會計 \\ 行銷管理 \\ 
    \end{tabular} &
    \begin{tabular}[c]{c}155 \\ 164 \\ 129 \\ 399 \\ 173 \\ 229 \\ 226 \\ 148 \\ 238 \\ 227 \\ 714 \\ 463 \\ 1148 \\ 118 \\ 197 \\ 106 \\ 
    \end{tabular} \\ 
    
    \midrule
    
    Humanities & 
    \begin{tabular}[c]{l}Chinese National Conditions \\ Chinese and World Culture \\ Chinese History \\ Hong Kong Education \# \\ Hong Kong Historical Geography and Development \# \\ Hong Kong History \# \\ Hong Kong Cultural Art \# \\ Hong Kong Current Affairs \# \\ Hong Kong Celebrity \# \\ Hong Kong Government and Political System \# \\ Hong Kong Literature \# \\ Hong Kong Politicians and Events \# \\ Hong Kong Society \# \\ Hong Kong Human Geography \# \\ Hong Kong Law \# \\ Hong Kong Party Politics \# \\ Human Behavior \\ World Entertainment \\ 
    \end{tabular} &
    \begin{tabular}[c]{l}中國國情 \\ 中國與世界文化 \\ 中國歷史 \\ 香港教育 \\ 香港歷史地理與發展 \\ 香港歷史 \\ 香港文娛 \\ 香港時事 \\ 香港名人文化 \\ 香港政府機構與政治制度 \\ 香港文學 \\ 香港政治人物與事件 \\ 香港社會 \\ 香港人文地理 \\ 香港法律 \\ 香港政黨政治 \\ 人類行為與社會學 \\ 世界文娛 \\ 
    \end{tabular} &
    \begin{tabular}[c]{c}117 \\ 154 \\ 110 \\ 730 \\ 143 \\ 556 \\ 162 \\ 278 \\ 1267 \\ 281 \\ 203 \\ 268 \\ 515 \\ 1251 \\ 412 \\ 1440 \\ 297 \\ 720 \\ 
    \end{tabular} \\ 
    
    \midrule
    
    Other & 
    \begin{tabular}[c]{l}Commonsense of Science and Technology \\ Cultural Commonsense \\ Fire Science \\ Hong Kong Events and Festivals \# \\ Hong Kong Entertainment Industry and Management \# \\ Hong Kong Film and Television \# \\ Hong Kong Life Knowledge \# \\ Hong Kong Livelihood \# \\ Hong Kong Music \# \\ Hong Kong Transportation \# \\ Logic Reasoning \\ Nautical Science \\ Social Commonsense \\ Technical Skills \\ 
    \end{tabular} &
    \begin{tabular}[c]{l}科學技術常識 \\ 文化常識 \\ 火災學 \\ 香港活動與節慶 \\ 香港文娛產業與經營 \\ 香港影視 \\ 香港生活常識 \\ 香港民生 \\ 香港音樂 \\ 香港交通 \\ 邏輯思維 \\ 航海 \\ 社會常識 \\ 技術工學 \\ 
    \end{tabular} &
    \begin{tabular}[c]{c}124 \\ 164 \\ 135 \\ 108 \\ 155 \\ 2201 \\ 330 \\ 895 \\ 1035 \\ 357 \\ 159 \\ 606 \\ 187 \\ 346 \\ 
    \end{tabular} \\ 
    
    \bottomrule

\end{tabular}
}
    \caption{Overview of subjects in HKMMLU. ``\#'' indicated the HK-Specific subjects.}
\label{tab:subject_details}

\end{CJK}

\end{table*}

\begin{table*}[ht]
    
    \small
    \centering
    \resizebox{1\linewidth}{!}{\begin{tabular}{lccccccc}
    \toprule
Category & Subject & Q & Avg. Q & Max. Q & Min. Q & Avg. Q Tokens & Avg. C Tokens \\
        \midrule
STEM            & 18 & 6158 & 342.1 & 1020 & 103 & 34.8 & 36.6 \\
Humanities      & 16 & 4834 & 302.1 & 1148 & 106 & 27.3 & 28.2 \\
Social Science  & 18 & 8904 & 494.6 & 1440 & 110 & 17.9 & 24.2 \\
Other           & 14 & 6802 & 485.9 & 2201 & 108 & 22.0 & 23.7 \\
\midrule
All             & 66 & 26698 & 404.5 & 2201 & 103 & 24.5 & 27.7 \\ 
\bottomrule
    \end{tabular}
    }
    \caption{The statistics of the HKMMLU, where Q represents the question and C indicates the answer choices.}
    \label{tab:statistics}
 
\end{table*}

Table~\ref{tab:subject_details} lists all 66 subjects across the four categories, their names in Traditional Chinese, and the number of questions. In particular, HKMMLU includes 23 Hong Kong-specific subjects that cover local geography, history, politics, and culture unique to Hong Kong. Meanwhile, Basic Medical Sciences, Hong Kong Urban and Regional Planning, Hong Kong Party Politics, and Hong Kong Film and Television are the subjects with the most questions in their respective disciplines. Furthermore, Hong Kong Film and Television has the most significant data volume among all subjects in HKMMLU, which has 2,201 questions.

Table~\ref{tab:statistics} displays a detailed statistical breakdown of the HKMMLU by category, including the number of subjects, total questions, average number of questions per category, maximum and minimum counts, and the average token length of both questions and choices.

\section{Models Evaluated in this Paper}
\begin{table*}[htbp]
  \centering
  \scriptsize
  \renewcommand{\arraystretch}{0.9}
  \resizebox{1\textwidth}{!}{
  \begin{tabular}{lcccl}
    \toprule
    \textbf{Model} & \textbf{Version} & \textbf{Model Size} & \textbf{Access} & \textbf{Creator} \\
    \midrule
    \texttt{DeepSeek-V3}                    & - & 685B & API & deepseek \\
    \texttt{GPT-4o}                         & 20240806 & undisclosed & API & OpenAI \\
    \texttt{Claude 3.7 Sonnet}              & 20250219 & undisclosed & API & Anthropic \\
    \texttt{Gemma-2-2B-IT}                  & - & 2.61B & weights & Google \\
    \texttt{Gemma-2-27B-IT}                 & - & 27.2B & weights & Google \\
    \texttt{GLM-4-9B-Chat}                  & - & 9.4B & weights & Tsinghua \& Zhipu \\
    \texttt{Llama-3.1-8B-Instruct}          & 20240723 & 8.03B & weights & Meta AI \\
    \texttt{Llama-3.1-70B-Instruct}         & 20240723 & 70.6B & weights & Meta AI \\
    \texttt{Llama-3-Taiwan-8B-Instruct}     & - & 8.03B & weights & Yen-Ting Lin \\
    \texttt{Llama-3-Taiwan-70B-Instruct}    & - & 70.6B & weights & Yen-Ting Lin \\
    \texttt{Llama-3-8B-Instruct}            & 20240418 & 8.03B & weights & Meta AI \\
    \texttt{Llama-3-70B-Instruct}           & 20240418 & 70.6B & weights & Meta AI \\
    \texttt{Mistral-Small-Instruct}         & 2409 & 22.2B & weights & Mistral AI \\
    \texttt{Mistral-Large-Instruct}         & 2411 & 123B & weights & Mistral AI \\
    \texttt{Qwen2.5-3B-Instruct}            & - & 3.09B & weights & Alibaba \\
    \texttt{Qwen2.5-7B-Instruct}            & - & 7.61B & weights & Alibaba \\
    \texttt{Qwen2.5-14B-Instruct}           & - & 14.7B & weights & Alibaba \\
    \texttt{Qwen2.5-72B-Instruct}           & - & 72.7B & weights & Alibaba \\
    \texttt{Yi-1.5-9B-Chat}                 & - & 8.83B & weights & 01.AI \\
    \texttt{Yi-1.5-34B-Chat}                & - & 34.4B & weights & 01.AI \\     

  \bottomrule
  \end{tabular}
  }
  \caption{\label{model-details} Models evaluated in this paper.}
\end{table*}

Table~\ref{model-details} outlines the models evaluated in this paper, including their versions, sizes, access methods, and creators.

\section{Detailed Results of Few-shot Prompting}

\begin{center}
\scriptsize

\begin{longtable}{lcccccc}
    
    \toprule
    \textbf{Models} & \textbf{Shot Num} & \textbf{STEM} & \textbf{Soc. Sci.} & \textbf{Humanities} & \textbf{Other} & \textbf{Avg.} \\

    \midrule
    \multirow{6}{*}{\texttt{DeepSeek-V3}} 
    & 0 & 74.5 & 73.5 & 76.0 & 74.8 & 74.9 \\
    & 1 & 74.0 & 73.5 & 77.4 & 74.6 & 75.2 \\
    & 2 & 74.0 & 73.2 & 77.3 & 74.8 & 75.1 \\
    & 3 & \textbf{75.0} & 73.8 & \textbf{77.6} & \textbf{76.2} & \textbf{75.9} \\
    & 4 & 74.7 & \textbf{73.9} & 77.5 & 75.5 & 75.7 \\
    & 5 & 74.9 & \textbf{73.9} & 77.3 & 75.5 & 75.6 \\

    \midrule
    \multirow{6}{*}{\texttt{GPT-4o}}
    & 0 & \textbf{75.0} & \textbf{70.4} & 68.6 & 67.9 & \textbf{70.3} \\
    & 1 & 74.7 & 68.7 & 68.4 & 66.6 & 69.5 \\
    & 2 & 74.3 & 69.1 & \textbf{69.4} & 67.9 & 70.1 \\
    & 3 & 74.0 & 69.2 & 68.8 & 67.2 & 69.7 \\
    & 4 & 74.2 & 68.9 & 69.1 & \textbf{68.0} & 70.0 \\
    & 5 & \textbf{75.0} & 68.9 & 69.2 & 67.3 & 70.0 \\
    
    \midrule
    \multirow{6}{*}{\texttt{Claude 3.7 Sonnet}}  
    & 0 & \textbf{74.3} & 69.4 & 63.0 & 62.1 & 66.7 \\
    & 1 & 74.1 & \textbf{69.8} & 68.3 & \textbf{62.8} & \textbf{68.7} \\
    & 2 & \textbf{74.3} & 69.2 & 66.7 & \textbf{62.8} & 68.1 \\
    & 3 & 74.2 & 68.4 & 63.8 & 62.1 & 66.8 \\
    & 4 & 57.7 & 67.3 & \textbf{68.8} & 59.1 & 63.7 \\
    & 5 & 46.3 & 49.6 & 48.4 & 45.2 & 47.4 \\
    
    \midrule
    \multirow{6}{*}{\texttt{Gemma-2-2B-IT}} 
    & 0 & \textbf{30.6} & \textbf{40.1} & 47.7 & 47.0 & 42.0 \\
    & 1 & 30.3 & 37.6 & 48.0 & 47.1 & 41.5 \\
    & 2 & 30.0 & 38.5 & 48.3 & 48.1 & 42.0 \\
    & 3 & 30.3 & 39.4 & \textbf{50.0} & 48.0 & \textbf{42.8} \\
    & 4 & 30.5 & 37.5 & 48.6 & \textbf{48.3} & 42.1 \\
    & 5 & 30.3 & 39.6 & 49.9 & 47.7 & \textbf{42.8} \\
    
    \midrule
    \multirow{6}{*}{\texttt{Gemma-2-27B-IT}}
    & 0 & \textbf{55.7} & 55.4 & 59.7 & 57.5 & 57.4 \\
    & 1 & 55.0 & 55.8 & 59.4 & 58.0 & 57.3 \\
    & 2 & 55.3 & \textbf{56.0} & 60.0 & 57.8 & 57.6 \\
    & 3 & 54.4 & 55.7 & 59.5 & \textbf{58.6} & 57.3 \\
    & 4 & 54.7 & 55.8 & 59.8 & 58.2 & 57.5 \\
    & 5 & 55.2 & \textbf{56.0} & \textbf{60.3} & 57.9 & \textbf{57.7} \\
    
    \midrule
    \multirow{6}{*}{\texttt{GLM-4-9B-Chat}} 
    & 0 & \textbf{42.8} & \textbf{47.3} & \textbf{53.4} & \textbf{49.9} & \textbf{48.9} \\
    & 1 & 38.8 & 43.8 & 50.9 & 47.5 & 45.9 \\
    & 2 & 40.2 & 46.1 & 50.8 & 48.9 & 46.9 \\
    & 3 & 39.1 & 45.9 & 51.1 & \textbf{49.9} & 47.0 \\
    & 4 & 39.9 & 46.2 & 51.9 & 49.3 & 47.4 \\
    & 5 & 39.2 & 46.1 & 51.3 & 48.7 & 46.9 \\
    
    \midrule
    \multirow{6}{*}{\texttt{Llama-3.1-8B-Instruct}} 
    & 0 & 32.6 & 38.1 & 44.4 & 42.7 & 40.0 \\
    & 1 & \textbf{35.2} & 38.1 & 42.9 & 42.8 & 40.1 \\
    & 2 & 34.1 & \textbf{39.2} & 44.3 & 41.3 & 40.2 \\
    & 3 & 33.9 & 38.6 & 44.5 & \textbf{43.6} & \textbf{40.6} \\
    & 4 & 33.0 & 38.0 & \textbf{45.4} & 43.5 & \textbf{40.6} \\
    & 5 & 32.6 & 37.7 & 44.1 & 43.0 & 39.9 \\
        
    \midrule
    \multirow{6}{*}{\texttt{Llama-3.1-70B-Instruct}} 
    & 0 & \textbf{60.5} & \textbf{59.5} & \textbf{58.8} & 55.9 & \textbf{58.7} \\
    & 1 & 59.7 & 57.3 & 58.5 & 56.7 & 58.1 \\
    & 2 & 59.8 & 58.1 & 58.5 & 55.9 & 58.1 \\
    & 3 & 59.6 & 58.2 & 57.7 & 56.9 & 58.1 \\
    & 4 & 59.6 & 57.9 & 58.6 & 56.1 & 58.1 \\
    & 5 & 59.8 & 56.9 & 58.3 & \textbf{57.1} & 58.1 \\
    
    \midrule
    \multirow{6}{*}{\texttt{Llama-3-Taiwan-8B-Instruct}} 
    & 0 & \textbf{47.4} & 48.9 & \textbf{56.6} & 52.4 & \textbf{51.9} \\
    & 1 & 46.2 & 48.1 & 54.8 & 51.6 & 50.7 \\
    & 2 & 45.9 & 48.1 & 54.8 & 52.1 & 50.7 \\
    & 3 & 46.4 & 47.9 & 56.5 & 52.5 & 51.5 \\
    & 4 & 46.1 & \textbf{49.6} & 55.6 & 53.1 & 51.6 \\
    & 5 & 45.5 & 49.0 & 56.3 & \textbf{53.4} & 51.6 \\
    
    \midrule
    \multirow{6}{*}{\texttt{Llama-3-Taiwan-70B-Instruct}} 
    & 0 & \textbf{73.6} & 67.3 & \textbf{66.3} & 62.6 & \textbf{67.4} \\
    & 1 & 72.4 & 67.9 & 65.1 & 63.0 & 66.9 \\
    & 2 & 72.9 & 67.8 & 65.6 & 63.7 & 67.3 \\
    & 3 & 71.8 & 67.0 & 64.5 & \textbf{65.0} & 66.8 \\
    & 4 & 72.3 & 66.6 & 64.8 & 63.2 & 66.6 \\
    & 5 & 73.1 & \textbf{68.0} & 64.9 & 64.0 & 67.2 \\
    
    \midrule
    \multirow{6}{*}{\texttt{Llama-3-8B-Instruct}} 
    & 0 & \textbf{33.3} & 41.3 & 53.2 & 48.6 & 45.1 \\
    & 1 & 32.9 & 35.7 & 46.1 & 43.8 & 40.4 \\
    & 2 & \textbf{33.3} & 41.0 & 52.8 & 50.0 & 45.2 \\
    & 3 & 32.2 & 41.2 & 53.2 & 49.8 & 45.1 \\
    & 4 & 32.4 & \textbf{42.0} & \textbf{53.3} & \textbf{50.1} & \textbf{45.4} \\
    & 5 & 29.9 & 40.5 & 49.1 & 45.6 & 42.1 \\
    
    \midrule
    \multirow{6}{*}{\texttt{Llama-3-70B-Instruct}} 
    & 0 & \textbf{59.7} & 57.8 & 61.0 & 57.2 & 59.2 \\
    & 1 & 58.2 & 56.8 & 61.2 & 59.2 & 59.1 \\
    & 2 & 57.6 & 57.7 & 61.7 & 59.1 & 59.4 \\
    & 3 & 57.6 & 57.6 & \textbf{62.2} & \textbf{60.5} & 59.8 \\
    & 4 & 57.6 & 57.0 & 62.1 & \textbf{60.5} & 59.6 \\
    & 5 & 58.1 & \textbf{58.2} & 61.9 & 60.1 & \textbf{59.9} \\

    \midrule
    \multirow{6}{*}{\texttt{Mistral-Small-Instruct}} 
    & 0 & 38.9 & 42.1 & 49.4 & 47.8 & 45.1 \\
    & 1 & 41.3 & \textbf{43.6} & 51.1 & \textbf{50.2} & 47.1 \\
    & 2 & 40.7 & 43.5 & 51.9 & 49.4 & 47.0 \\
    & 3 & \textbf{41.8} & 42.4 & 52.8 & 48.6 & 47.2 \\
    & 4 & 41.3 & 43.3 & 52.5 & 48.6 & 47.1 \\
    & 5 & 41.7 & 43.4 & \textbf{53.3} & 49.0 & \textbf{47.6} \\

    \midrule
    \multirow{6}{*}{\texttt{Mistral-Large-Instruct}} 
    & 0 & 63.8 & 58.2 & 59.5 & 58.4 & 60.0 \\
    & 1 & 64.0 & 58.1 & 59.3 & \textbf{58.8} & 60.0 \\
    & 2 & 63.5 & \textbf{58.6} & \textbf{60.0} & 58.3 & 60.1 \\
    & 3 & 63.6 & 58.1 & 59.4 & 58.6 & 59.9 \\
    & 4 & \textbf{64.6} & 58.5 & 59.6 & 58.1 & \textbf{60.2} \\
    & 5 & 63.9 & 57.8 & 59.0 & 58.1 & 59.6 \\
    
    \midrule
    \multirow{6}{*}{\texttt{Qwen2.5-3B-Instruct}} 
    & 0 & \textbf{42.5} & 48.0 & 54.1 & 54.7 & 50.3 \\
    & 1 & 42.3 & 47.3 & 53.2 & 54.5 & 49.8 \\
    & 2 & \textbf{42.5} & 48.0 & 54.2 & 54.8 & 50.3 \\
    & 3 & 42.0 & 47.7 & \textbf{55.6} & \textbf{56.4} & \textbf{51.0} \\
    & 4 & 42.0 & \textbf{48.1} & 55.3 & 55.4 & 50.7 \\
    & 5 & 42.1 & 47.5 & 54.7 & 54.3 & 50.2 \\
        
    \midrule
    \multirow{6}{*}{\texttt{Qwen2.5-7B-Instruct}} 
    & 0 & 56.0 & 54.1 & 60.0 & 57.4 & 57.2 \\
    & 1 & 55.9 & 54.1 & 59.6 & 57.1 & 57.0 \\
    & 2 & 56.0 & 54.2 & 59.3 & 56.9 & 56.9 \\
    & 3 & \textbf{56.6} & 55.0 & 60.2 & 57.3 & 57.6 \\
    & 4 & 56.2 & 55.3 & \textbf{61.3} & \textbf{57.9} & \textbf{58.1} \\
    & 5 & 56.1 & \textbf{55.7} & 60.6 & 57.1 & 57.7 \\
    
    \midrule
    \multirow{6}{*}{\texttt{Qwen2.5-14B-Instruct}} 
    & 0 & 63.9 & 61.5 & 62.8 & 60.4 & 62.3 \\
    & 1 & 63.0 & 61.6 & 63.8 & \textbf{62.6} & 62.9 \\
    & 2 & 63.5 & 61.7 & 63.1 & 62.3 & 62.7 \\
    & 3 & 63.5 & \textbf{61.9} & 64.2 & 61.7 & 63.0 \\
    & 4 & 63.7 & 61.5 & 65.0 & 61.5 & 63.2 \\
    & 5 & \textbf{64.1} & 61.8 & \textbf{65.1} & 61.8 & \textbf{63.4} \\

    \midrule
    \multirow{6}{*}{\texttt{Qwen2.5-72B-Instruct}} 
    & 0 & 72.0 & \textbf{67.9} & \textbf{70.1} & \textbf{65.8} & \textbf{69.1} \\
    & 1 & 71.2 & 66.2 & 68.4 & 64.7 & 67.7 \\
    & 2 & \textbf{72.5} & 66.8 & 68.9 & 64.6 & 68.3 \\
    & 3 & 72.3 & 66.7 & 68.4 & 65.1 & 68.2 \\
    & 4 & 72.1 & 66.5 & 68.9 & 65.3 & 68.3 \\
    & 5 & 71.9 & 67.2 & 68.7 & 64.8 & 68.3 \\
    
    \midrule
    \multirow{6}{*}{\texttt{Yi-1.5-9B-Chat}} 
    & 0 & 45.4 & 51.6 & 56.9 & 54.3 & 52.5 \\
    & 1 & 44.9 & 53.0 & 58.3 & 56.0 & 53.6 \\
    & 2 & 46.1 & 53.3 & 59.4 & 58.6 & 54.9 \\
    & 3 & 46.8 & 54.2 & \textbf{60.9} & 57.9 & 55.6 \\
    & 4 & 46.5 & 54.0 & 59.9 & 58.1 & 55.2 \\
    & 5 & \textbf{47.0} & \textbf{54.6} & 60.8 & \textbf{58.8} & \textbf{55.9} \\
    
    \midrule
    \multirow{6}{*}{\texttt{Yi-1.5-34B-Chat}} 
    & 0 & 56.2 & 58.7 & 64.6 & 60.5 & 60.5 \\
    & 1 & 54.6 & 59.7 & 65.3 & 61.2 & 60.7 \\
    & 2 & 56.6 & 59.1 & 66.3 & 61.2 & 61.4 \\
    & 3 & \textbf{56.8} & 59.5 & 66.3 & \textbf{61.8} & 61.7 \\
    & 4 & 56.1 & \textbf{59.8} & \textbf{66.8} & \textbf{61.8} & \textbf{61.8} \\
    & 5 & 56.6 & 59.5 & 65.9 & 60.8 & 61.3 \\

    \bottomrule

\caption{
    \label{appendix-few-shot}
    Accuracy (\%) in multi-choice tasks with different models, showing 0-shot to 5-shot performance for each model.
    }

\end{longtable}
\end{center}

We compared the detailed results of few-shot prompting on HKMMLU across four categories in Table~\ref{appendix-few-shot}. We found that increasing the number of shots for a single model does not necessarily improve its performance, underscoring the importance of knowledge acquisition during the pre-training process. Specifically, Claude 3.7 Sonnet exhibits a sharp decline in performance from 4-shot to 5-shot, with many responses being, ``Sorry, I didn't understand your query. Can you provide more details?'' This decline indicates that adding more examples does not enhance performance but may lead to confusion.

\section{Efficiency of Reasoning Examples for STEM and non-STEM Tasks}

\begin{table*}[h]
  \centering
  \fontsize{8.5pt}{8.5pt}\selectfont 
    \setlength{\tabcolsep}{3.4pt}
    \renewcommand{\arraystretch}{0.9}
    \resizebox{0.8\textwidth}{!}{
  \begin{tabular}{l|ccc|ccc}
    \toprule
    \multirow{2}{*}{\textbf{Models}}  & \multicolumn{3}{c|}{\textbf{STEM}} & \multicolumn{3}{c}{\textbf{Non-STEM}} \\
     & DA~~&~1s~~&~1s-cot & DA~~&~1s~~&~1s-cot \\
    \midrule
    
    \texttt{DeepSeek-V3} & 74.5 & 74.0 & \textbf{81.0} & 75.0 & \textbf{75.5} & 74.2 \\
    \texttt{GPT-4o} & 75.0 & 74.7 & \textbf{82.1} & 68.9 & 68.0 & \textbf{69.3} \\
    \texttt{Claude 3.7 Sonnet} & 74.3 & 74.1 & \textbf{78.6} & 64.5 & \textbf{67.1} & 63.3 \\
    \texttt{Gemma-2-2B-IT} & \textbf{30.6} & 30.3 & 30.2 & \textbf{45.4} & 44.9 & 38.8 \\
    \texttt{Gemma-2-27B-IT} & 55.7 & 55.0 & \textbf{60.6} & 57.9 & \textbf{58.0} & 55.8 \\
    \texttt{GLM-4-9B-Chat} & 42.8 & 38.8 & \textbf{51.2} & 50.7 & 48.0 & \textbf{51.0} \\
    \texttt{Llama-3.1-8B-Instruct} & 32.6 & \textbf{35.2} & 29.0 & \textbf{42.2} & 41.6 & 30.0 \\
    \texttt{Llama-3.1-70B-Instruct} & \textbf{60.5} & 59.7 & 43.2 & \textbf{58.1} & 57.7 & 38.5 \\
    \texttt{Llama-3-Taiwan-8B-Instruct} & 47.4 & 46.2 & \textbf{48.3} & \textbf{53.3} & 52.0 & 52.0 \\
    \texttt{Llama-3-Taiwan-70B-Instruct} & 73.6 & 72.4 & \textbf{74.4} & \textbf{65.5} & 65.2 & 64.1 \\
    \texttt{Llama-3-8B-Instruct} & 33.3 & 32.9 & \textbf{37.6} & \textbf{48.6} & 42.6 & 48.2 \\
    \texttt{Llama-3-70B-Instruct} & 59.7 & 58.2 & \textbf{62.4} & 59.0 & 59.4 & \textbf{59.9} \\
    \texttt{Mistral-Small-Instruct} & 38.9 & \textbf{41.3} & 40.2 & 47.0 & \textbf{48.8} & 40.2 \\
    \texttt{Mistral-Large-Instruct} & 63.8 & \textbf{64.0} & 51.7 & \textbf{58.8} & \textbf{58.8} & 48.6 \\
    \texttt{Qwen2.5-3B-Instruct} & \textbf{42.5} & 42.3 & 41.8 & \textbf{52.6} & 52.0 & 44.7 \\
    \texttt{Qwen2.5-7B-Instruct} & 56.0 & 55.9 & \textbf{56.3} & \textbf{57.6} & 57.4 & 53.2 \\
    \texttt{Qwen2.5-14B-Instruct} & 63.9 & 63.0 & \textbf{64.7} & 61.8 & \textbf{62.9} & 60.8 \\
    \texttt{Qwen2.5-72B-Instruct} & \textbf{72.0} & 71.2 & 71.1 & \textbf{68.3} & 66.7 & 65.2 \\
    \texttt{Yi-1.5-9B-Chat} & \textbf{45.4} & 44.9 & 45.4 & 54.7 & \textbf{56.2} & 52.0 \\
    \texttt{Yi-1.5-34B-Chat} & \textbf{56.2} & 54.6 & 55.5 & 61.8 & \textbf{62.6} & 58.9 \\
                       
    \midrule
    Avg. & 54.9 & 54.4 & \textbf{55.3} & \textbf{57.6} & 57.3 & 53.4 \\

    \bottomrule
  \end{tabular}
  }
  \caption{\label{compare_stem_non_stem} 
    Comparison results on DA, 1-shot and 1-shot-CoT.
  }
\end{table*}

As shown in Table~\ref{compare_stem_non_stem}, in the STEM category, models such as DeepSeek-V3, GPT-4o, Gemma-2-27B-IT, GLM-4-9B-Chat, the Llama-3 Taiwan series, the Llama-3 series, Qwen-2.5-3B/14B-Instruct, and the Yi-1.5 series demonstrated a decline in performance after applying a 1-shot example. However, their performance significantly improved after incorporating the reasoning process, surpassing the results of 0-shot. In contrast, this enhancement is rare in non-STEM tasks, with only GPT-4o and GLM-4-9B-Chat exhibiting slight improvements from the 1-shot CoT approach.

This trend may be because the STEM category primarily includes subjects such as Mathematics and Physics, which require a reasoning process. However, instructing the LLM to think step-by-step does not guarantee a high-quality reasoning chain. Providing high-quality reasoning examples enables the LLM to organize its thought process more effectively, leading to improved performance. This finding suggests that high-quality reasoning examples significantly contribute to the accuracy of tasks that require systematic reasoning steps. 

\section{Simplified Chinese vs. Languages of Hong Kong}

\begin{CJK}{UTF8}{bsmi}

Simplified Chinese and Hong Kong languages share fundamental grammar, professional terminology, and basic vocabulary but differ significantly in character forms, vocabulary choices, grammatical habits, and cultural context. 
For example, simplified Chinese uses streamlined characters (e.g., ``体'' vs. Traditional ``體''), while Hong Kong has specific terms like ``MTR'' (港鐵) that differ from mainland equivalents. 
Furthermore, Cantonese often omits subjects and employs colloquial particles, requiring flexible grammatical parsing.
Therefore, cultural references and social issues unique to Hong Kong necessitate cultural knowledge for accurate interpretation. 
To perform well on Hong Kong-related tasks in the Hong Kong language, LLMs should incorporate region-specific vocabularies, flexible grammatical parsing, and cultural knowledge bases to enhance understanding across linguistic and cultural divides.

\end{CJK}

\section{Model Performance Comparison: CMMLU, TMMLU+, and HKMMLU}

\begin{table*}[h]
    \centering
    \resizebox{\columnwidth}{!}{
        \begin{tabular}{l|cccc|cccc|cccc}
            \toprule
            \multirow{2}{*}{\textbf{Model}}  & \multicolumn{4}{c|}{\textbf{HKMMLU}} & \multicolumn{4}{c|}{\textbf{CMMLU}} & \multicolumn{4}{c}{\textbf{TMMLU+}} \\
              & STEM & Soc.Sci & Humanities & Other & STEM & Soc.Sci & Humanities & Other & STEM & Soc.Sci & Humanities & Other \\
            \midrule
            \texttt{DeepSeek-V3} & \underline{74.5} & \textbf{73.5} & \textbf{76.0} & \textbf{74.8} & \underline{81.6} & \textbf{83.0} & \underline{85.1} & \underline{85.5} & \textbf{79.9} & \underline{81.5} & 68.5 & \underline{71.7} \\
            \texttt{GPT-4o} & \textbf{75.0} & \underline{70.4} & 68.6 & \underline{67.9} & 73.4 & 77.0 & 78.0 & 81.3 & \underline{76.1} & 80.7 & 62.5 & 70.7 \\
            \texttt{Claude 3.7 Sonnet} & 74.3 & 69.4 & 63.0 & 62.1 & 72.7 & 67.9 & 74.4 & 76.6 & 74.7 & 81.3 & \underline{69.2} & 70.3 \\
            \texttt{Gemma-2-2B-IT} & 30.6 & 40.1 & 47.7 & 47.0 & 37.3 & 43.5 & 40.4 & 48.7 & 32.3 & 36.1 & 29.5 & 33.4 \\
            \texttt{Gemma-2-27B-IT} & 55.7 & 55.4 & 59.7 & 57.5 & 58.7 & 63.0 & 62.0 & 68.9 & 57.3 & 63.1 & 45.6 & 53.0 \\
            \texttt{GLM-4-9B-Chat} & 42.8 & 47.3 & 53.4 & 49.9 & 59.3 & 69.6 & 71.2 & 68.8 & 47.3 & 52.9 & 39.9 & 44.7 \\
            \texttt{Llama-3.1-8B-Instruct} & 32.6 & 38.1 & 44.4 & 42.7 & 39.4 & 45.6 & 45.8 & 54.0 & 35.0 & 37.3 & 29.4 & 35.0 \\
            \texttt{Llama-3.1-70B-Instruct} & 60.5 & 59.5 & 58.8 & 55.9 & 64.3 & 66.6 & 66.2 & 73.1 & 62.5 & 65.2 & 47.9 & 54.4 \\
            \texttt{Llama-3-Taiwan-8B-Instruct} & 47.4 & 48.9 & 56.6 & 52.4 & 42.4 & 49.4 & 49.7 & 57.4 & 50.3 & 59.1 & 48.8 & 50.6 \\
            \texttt{Llama-3-Taiwan-70B-Instruct} & 73.6 & 67.3 & 66.3 & 62.6 & 61.1 & 66.2 & 67.8 & 72.3 & 75.4 & \textbf{83.9} & \textbf{73.8} & \textbf{74.1} \\
            \texttt{Llama-3-8B-Instruct} & 33.3 & 41.3 & 53.2 & 48.6 & 41.8 & 47.4 & 47.6 & 54.4 & 34.8 & 37.9 & 30.7 & 35.8 \\
            \texttt{Llama-3-70B-Instruct} & 59.7 & 57.8 & 61.0 & 57.2 & 63.0 & 65.0 & 64.5 & 73.0 & 61.7 & 63.0 & 49.0 & 55.0 \\
            \texttt{Mistral-Small-Instruct} & 38.9 & 42.1 & 49.4 & 47.8 & 43.4 & 50.6 & 48.8 & 56.1 & 39.3 & 43.1 & 31.5 & 37.9 \\
            \texttt{Mistral-Large-Instruct} & 63.8 & 58.2 & 59.5 & 58.4 & 64.6 & 69.5 & 70.8 & 72.0 & 65.4 & 68.2 & 50.6 & 57.6 \\
            \texttt{Qwen2.5-3B-Instruct} & 42.5 & 48.0 & 54.1 & 54.7 & 61.6 & 68.5 & 70.1 & 74.8 & 45.1 & 49.1 & 37.3 & 42.7 \\
            \texttt{Qwen2.5-7B-Instruct} & 56.0 & 54.1 & 60.0 & 57.4 & 75.0 & 77.8 & 80.9 & 80.6 & 58.5 & 62.0 & 48.8 & 55.2 \\
            \texttt{Qwen2.5-14B-Instruct} & 63.9 & 61.5 & 62.8 & 60.4 & 78.6 & 81.2 & 84.0 & 82.8 & 67.5 & 71.1 & 53.8 & 61.0 \\
            \texttt{Qwen2.5-72B-Instruct} & 72.0 & 67.9 & \underline{70.1} & 65.8 & \textbf{82.0} & \underline{82.7} & \textbf{85.8} & \textbf{85.7} & 74.4 & 78.1 & 60.8 & 67.8 \\
            \texttt{Yi-1.5-9B-Chat} & 45.4 & 51.6 & 56.9 & 54.3 & 64.9 & 72.4 & 73.7 & 75.8 & 49.3 & 56.6 & 43.7 & 47.5 \\
            \texttt{Yi-1.5-34B-Chat} & 56.2 & 58.7 & 64.6 & 60.5 & 70.3 & 74.5 & 77.4 & 77.3 & 55.7 & 62.3 & 46.2 & 52.4 \\                  
            \midrule
            Avg. & 54.9 & 55.6 & 59.3 & 56.9 & 61.8 & 66.1 & 67.2 & 71.0 & 57.1 & 61.6 & 48.4 & 53.5 \\
        
            \bottomrule
        \end{tabular}
    }
    \caption{\label{tab:detailed_mmlu_comp} 
        Model performance comparison on different MMLU Benchmark.
    }
\end{table*}

Table~\ref{tab:detailed_mmlu_comp} compares model performance on HKMMLU, CMMLU, and TMMLU+. These results indicate that most models perform the best on CMMLU in STEM, Social Sciences, and Humanities, while their performance is weaker on HKMMLU in STEM and Social Sciences. This discrepancy indicates a deficiency in LLMs' Traditional Chinese understanding abilities and highlights the challenges presented in our benchmark.

\section{Model Performance on Hong Kong-specific Tasks}

\begin{table*}[!t]
    \centering
    \footnotesize
    \setlength{\tabcolsep}{4.5pt}
    \renewcommand{\arraystretch}{0.9}
    \begin{tabular}{l|cc}
    
    \toprule

    \multirow{2}{*}{\textbf{Model}} & \multicolumn{2}{c}{\textbf{HK-Specific}} \\
    & tc~~&~sc  \\

    \midrule

    \texttt{DeepSeek-V3} & \textbf{75.1} & 74.6 \\
    \texttt{GPT-4o} & 67.8 & \textbf{68.1} \\
    \texttt{Claude 3.7 Sonnet} & 61.9 & \textbf{64.1}  \\
    \texttt{Gemma-2-2B-IT} & \textbf{48.0} & 46.8 \\
    \texttt{Gemma-2-27B-IT} & \textbf{57.7} & 57.1 \\
    \texttt{GLM-4-9B-Chat} & 51.4 & \textbf{53.0} \\
    \texttt{Llama-3.1-8B-Instruct} & 43.9 & \textbf{46.0} \\
    \texttt{Llama-3.1-70B-Instruct} & \textbf{57.5} & 54.9 \\
    \texttt{Llama-3-Taiwan-8B-Instruct} & \textbf{54.2} & 49.9 \\
    \texttt{Llama-3-Taiwan-70B-Instruct} & \textbf{63.7} & 61.7 \\
    \texttt{Llama-3-8B-Instruct} & \textbf{51.4} & 49.8 \\
    \texttt{Llama-3-70B-Instruct} & 59.3 & 59.4 \\
    \texttt{Mistral-Small-Instruct} & \textbf{48.4} & 47.6 \\
    \texttt{Mistral-Large-Instruct} & \textbf{57.8} & 56.9 \\
    \texttt{Qwen2.5-3B-Instruct} & \textbf{53.6} & 53.6 \\
    \texttt{Qwen2.5-7B-Instruct} & 57.3 & 57.4 \\
    \texttt{Qwen2.5-14B-Instruct} & 60.6 & \textbf{63.4} \\
    \texttt{Qwen2.5-72B-Instruct} & \textbf{67.0} & 66.9 \\
    \texttt{Yi-1.5-9B-Chat} & \textbf{55.4} & 54.4 \\
    \texttt{Yi-1.5-34B-Chat} & \textbf{62.4} & 62.1 \\

    \midrule
    Avg. & \textbf{57.7} & 57.4 \\

    \bottomrule
    \end{tabular}
    
    \caption{0-shot results of models on Hong Kong-specific (HK-specific) subject in HKMMLU.} 
    \label{tab:hk_specific}
\end{table*}

Table~\ref{tab:hk_specific} presents the performance of models on 23 subjects closely related to Hong Kong-specific knowledge. DeepSeek-V3 demonstrates superior performance in both Traditional Chinese and Simplified Chinese versions. Additionally, the Llama-3 Taiwan series does not exhibit the same advantages as the whole benchmark when focusing on Hong Kong-specific questions. This discrepancy may be due to their enhanced language capability in Traditional Chinese rather than an improvement in knowledge specific to Hong Kong.

\section{Model Performance by Subject}

\begin{figure}
    \centering
    \includegraphics[width=1.1\linewidth]{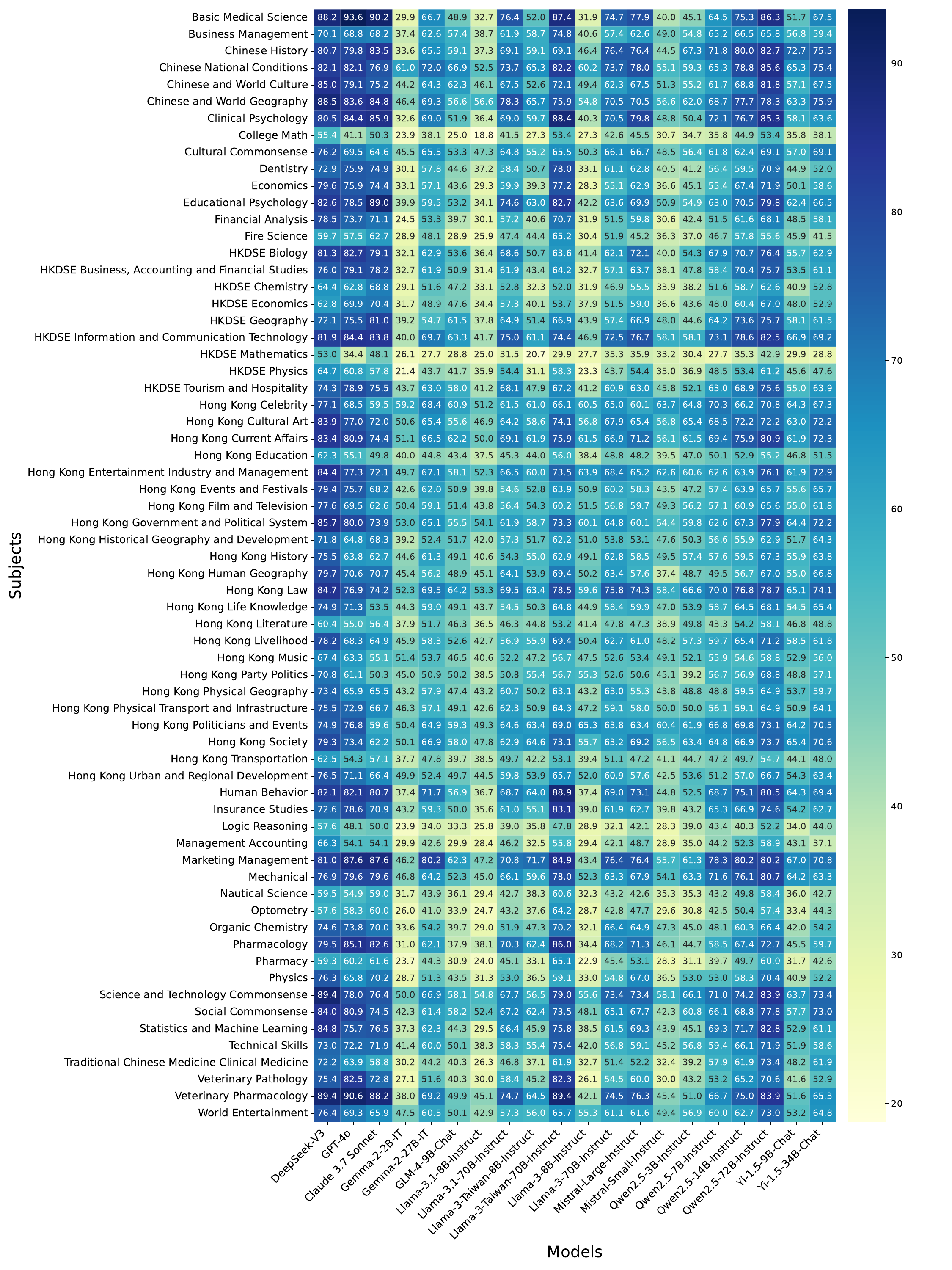}
    \caption{0-shot accuracy for 20 models across all subjects in HKMMLU}
    \label{fig:heatmap_0s_subject-models}
\end{figure}

\begin{figure}
    \centering
    \includegraphics[width=1.1\linewidth]{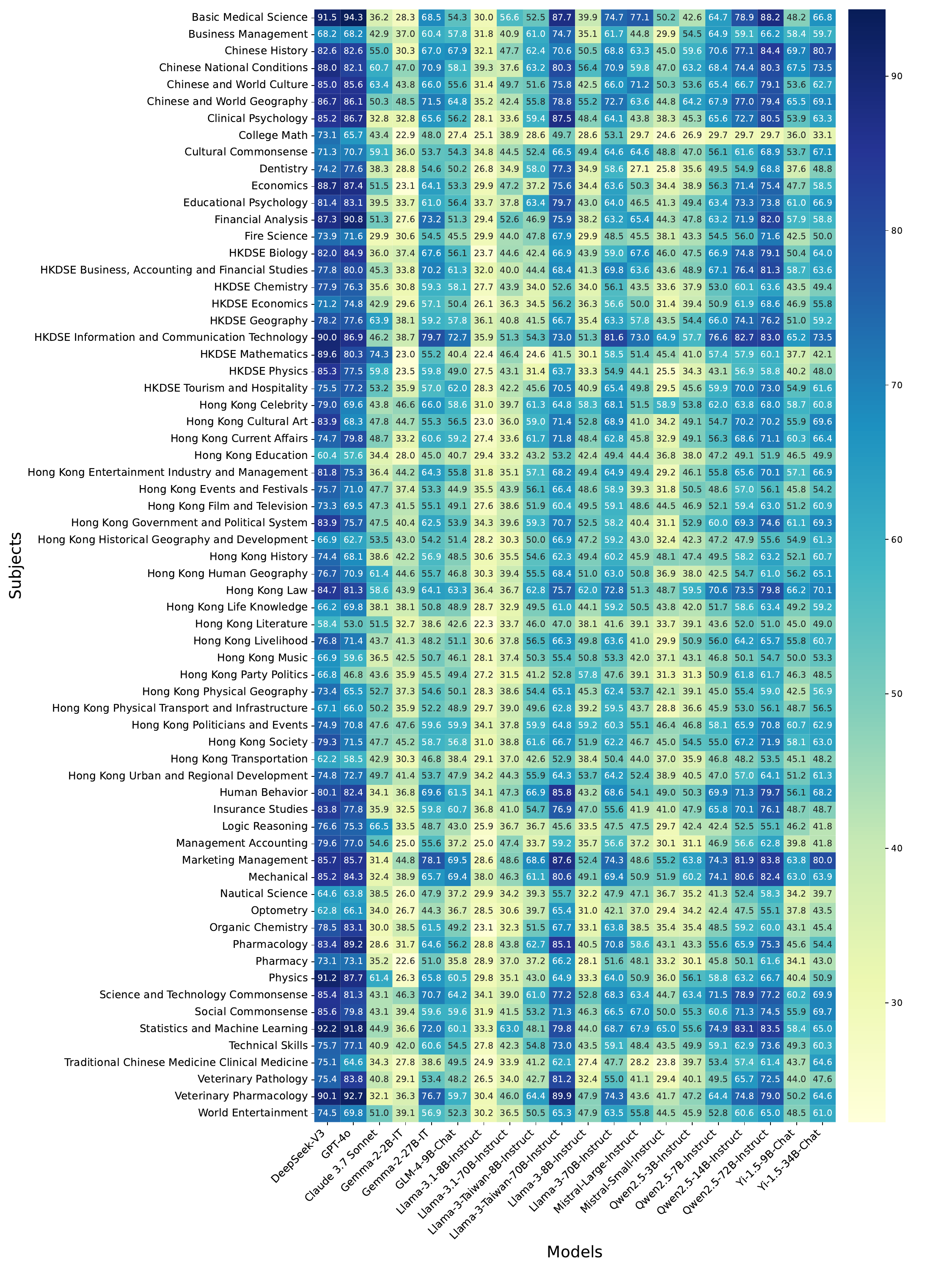}
    \caption{1-shot CoT accuracy for 20 models across all subjects in HKMMLU}
    \label{fig:heatmap_1cot_subject-models}
\end{figure}

Figures~\ref{fig:heatmap_0s_subject-models} and Figure~\ref{fig:heatmap_1cot_subject-models} respectively present the 0-shot and 1-shot CoT results for various models across all subjects in HKMMLU.

\section{Details of Human Review}
\begin{table*}[ht]
    \centering
    \begin{tabular}{lccc}
    \toprule
    Model & Accuracy Rate & Num. UQ & Num. UC  \\
    \midrule
    DeepSeek-V3         & 96 & 0 & 4 \\
    GPT-4o              & 97 & 1 & 2 \\
    Claude 3.7 Sonnet   & 99 & 1 & 0 \\
    \midrule
    Avg.                & 97.33 & 0.67 & 2 \\ 
    \bottomrule
    \end{tabular}
    \caption{The quality details of multi-choice that converted from question-answer pairs in HKMMLU, where ``Num. UQ'' represents the number of unreasonable questions and ``Num. UC'' indicates the number of unreasonable choices.}
    \label{tab:qa2mc}
 
\end{table*}
To ensure the quality of HKMMLU, we manually reviewed the multi-choice questions processed by the LLMs, including DeepSeek-V3, GPT-4o, and Claude 3.7 Sonnet. For each model, we manually checked 100 randomly selected questions. We assess the reasonableness of the question and answer choices. Results are displayed in Table~\ref{tab:qa2mc}. Additionally, to ensure the safety and ethical standards of HKMMLU, we have manually filtered out sensitive words.

\section{Prompts}

\subsection{Inference Prompts} \label{app:inference}

\begin{figure}[h]
\begin{center}
\includegraphics[width=1\linewidth]{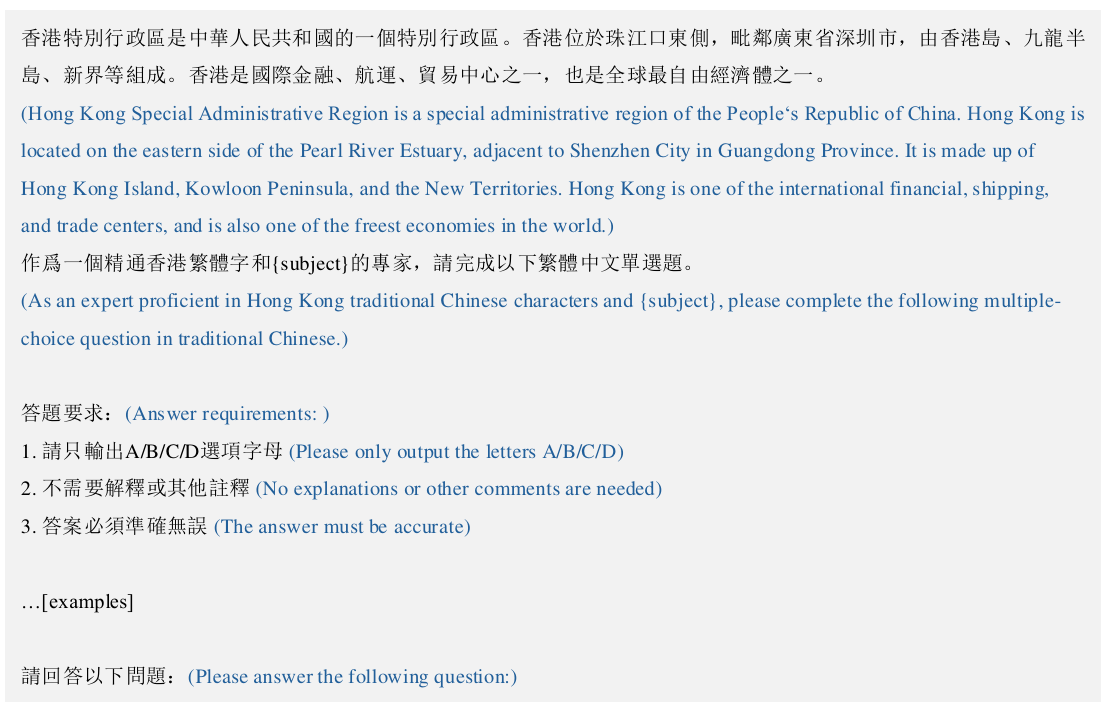} 
\end{center}
\caption{Few-shot prompt for multi-choice tasks in Traditional Chinese. 0-shot prompting involves no examples. English sentences in \textcolor{blue}{blue} are not part of the inference process.}
\label{fig:prompt-few-shot}
\end{figure}

\begin{figure}[h]
\begin{center}
\includegraphics[width=1\linewidth]{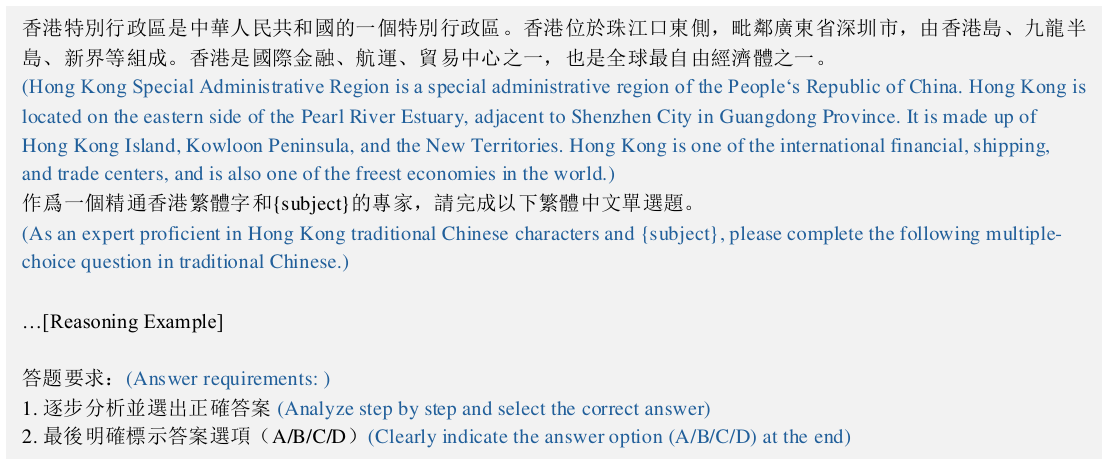} 

\end{center}
\caption{CoT prompt for multi-choice tasks in Traditional Chinese. 0-shot CoT prompting involves no examples. English sentences in \textcolor{blue}{blue} are not part of the inference process.}
\label{fig:prompt-cot}
\end{figure}

\begin{figure}[h]
\begin{center}
\includegraphics[width=1\linewidth]{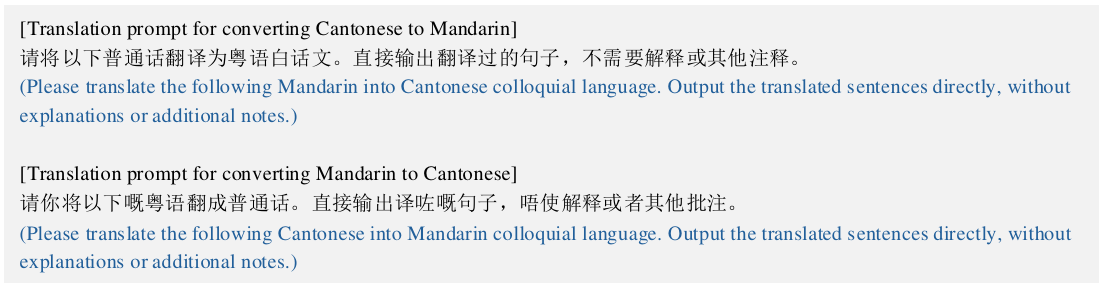} 
\end{center}
\caption{Prompt for translation tasks. English sentences in \textcolor{blue}{blue} are not part of the inference process.}
\label{fig:prompt-translation-tasks}
\end{figure}

Inference prompt for few-shot prompting is shown in Figure~\ref{fig:prompt-few-shot}, for CoT prompting in Figure~\ref{fig:prompt-cot}, and for translation tasks in Figure~\ref{fig:prompt-translation-tasks}.

\subsection{Translation Prompts}

\begin{figure}[h]
\begin{center}
\includegraphics[width=1\linewidth]{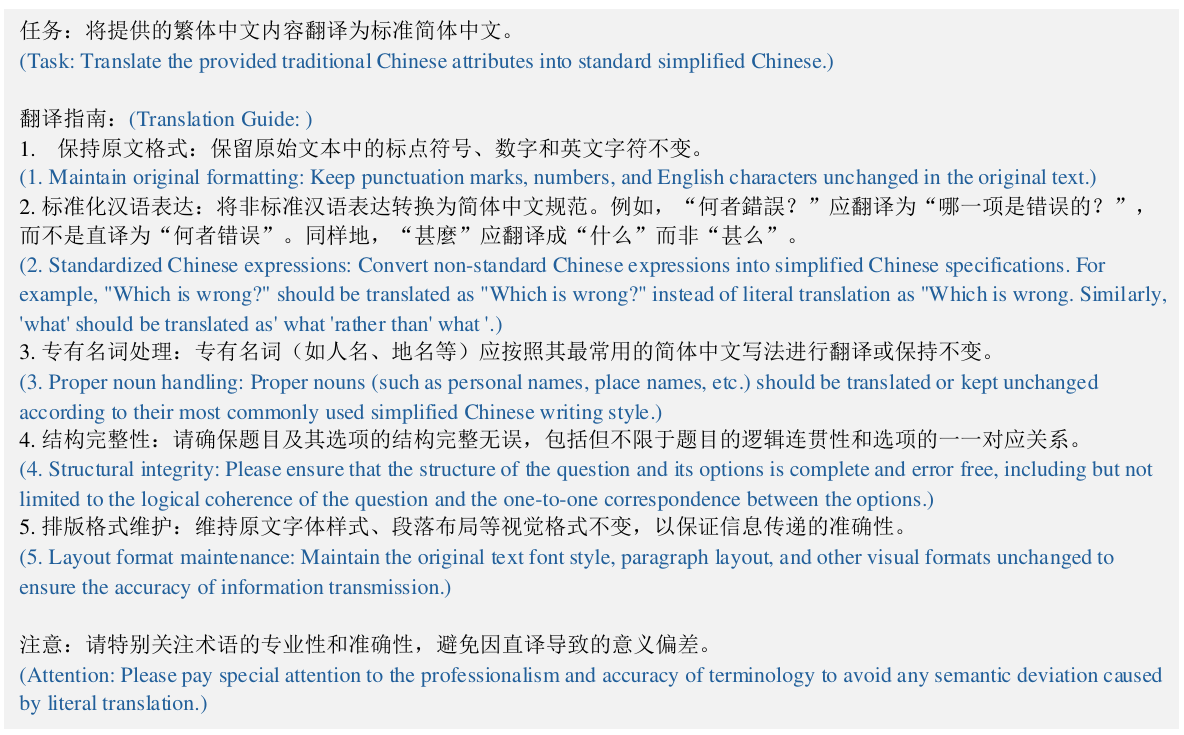} 
\end{center}
\caption{Prompt of translation from Traditional Chinese to Simplified Chinese. English sentences in \textcolor{blue}{blue} are not part of the inference process.}
\label{fig:prompt-translation}
\end{figure}

We use the prompt shown in Figure~\ref{fig:prompt-translation} to translate our multi-choice questions into their Simplified Chinese version.

\subsection{Subject Category Prompts}
\begin{figure}[h]
\begin{center}
\includegraphics[width=1\linewidth]{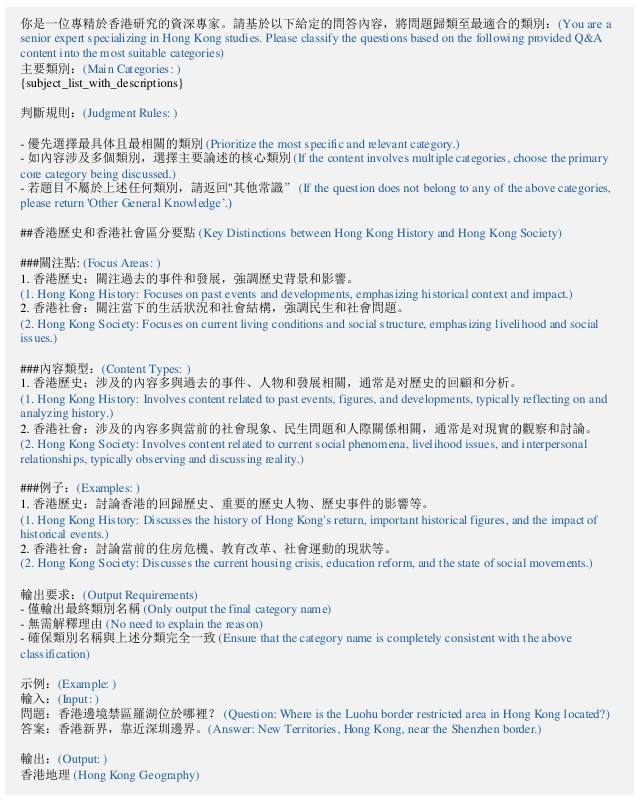} 
\end{center}
\caption{Prompt for categorizing questions into different subjects. English sentences in \textcolor{blue}{blue} are not part of the inference process.}
\label{fig:prompt-label}
\end{figure}

We use the prompt shown in Figure~\ref{fig:prompt-label} to categorize multi-choice questions.

\begin{figure}[h]
\begin{center}
\includegraphics[width=1\linewidth]{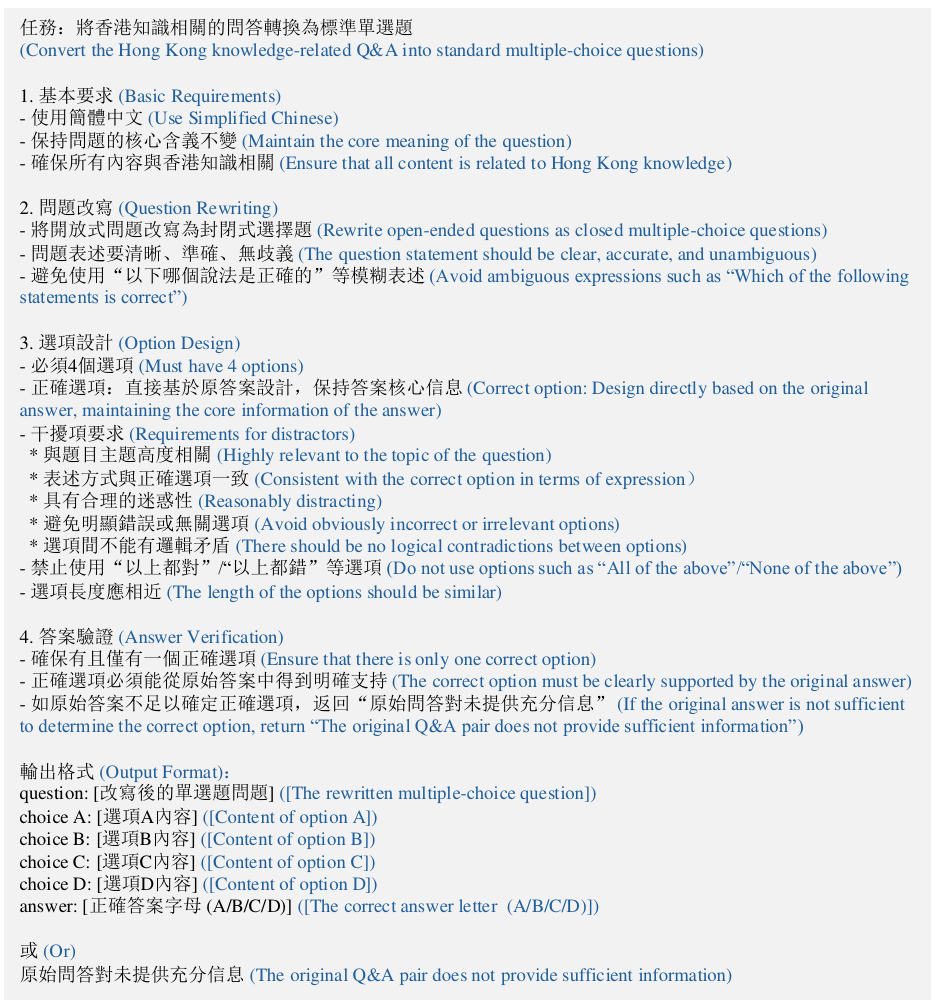} 
\end{center}
\caption{System prompt for processing question-answering pairs into multi-choice questions. English sentences in \textcolor{blue}{blue} are not part of the inference process.}
\label{fig:prompt-mc2qa}
\end{figure}

\section{Data Examples in HKMMLU}

\begin{figure}
    \centering
    \includegraphics[width=1\linewidth]{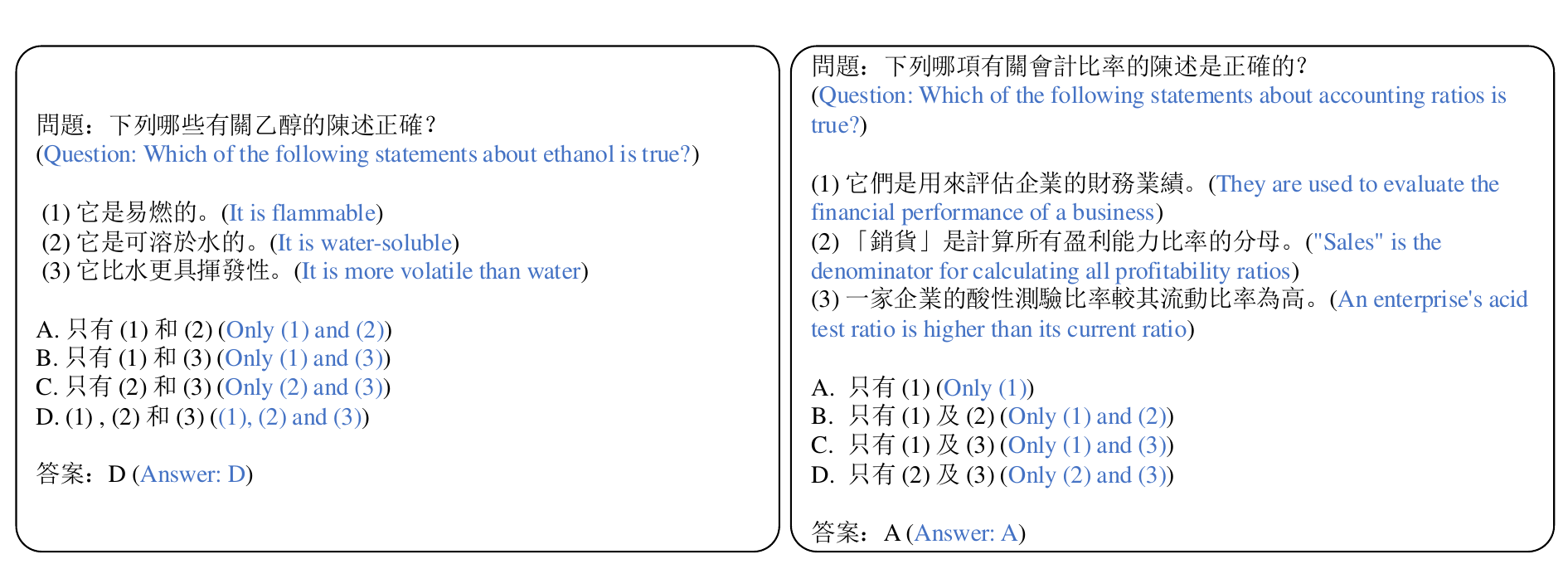}
    \caption{Examples of ``HKDSE Chemistry'' in STEM and ``HKDSE Business, Accounting, and Financial Studies'' in Social Science. The \textcolor{blue}{blue} text represents the English translation of the corresponding Chinese.}
    \label{fig:data_example_stem_socsci}
\end{figure}

\begin{figure}
    \centering
    \includegraphics[width=1\linewidth]{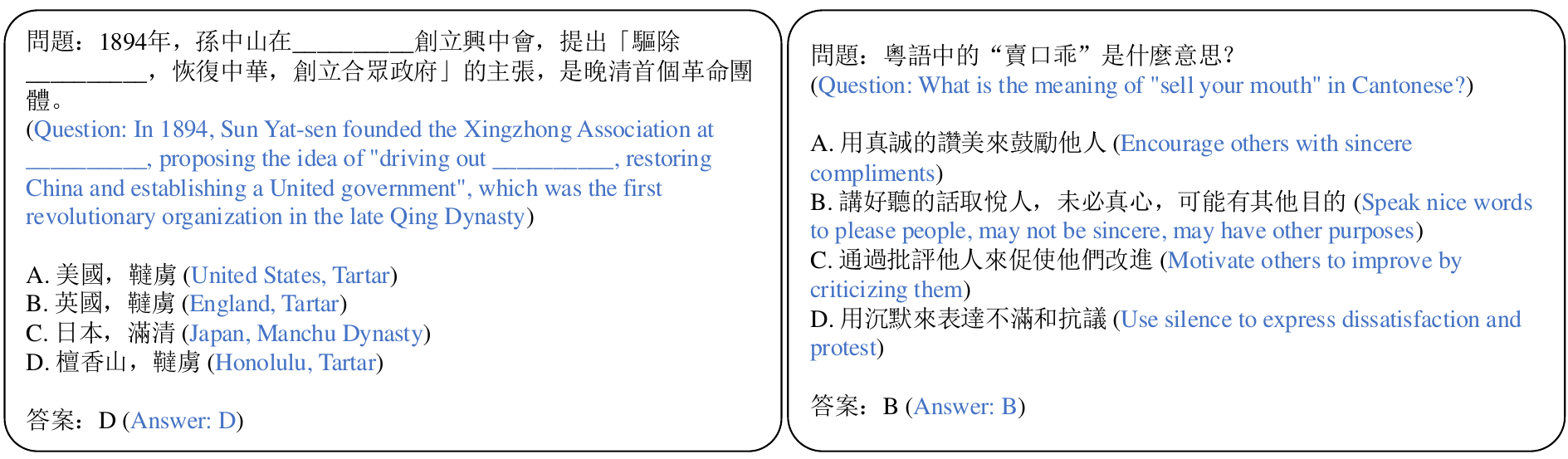}
    \caption{Examples of ``Chinese History'' in Humanities and ``Cultural Common Sense'' in Other. The \textcolor{blue}{blue} text represents the English translation of the corresponding Chinese.}
    \label{fig:data_example_human_other}
\end{figure}

\begin{figure}
    \centering
    \includegraphics[width=1\linewidth]{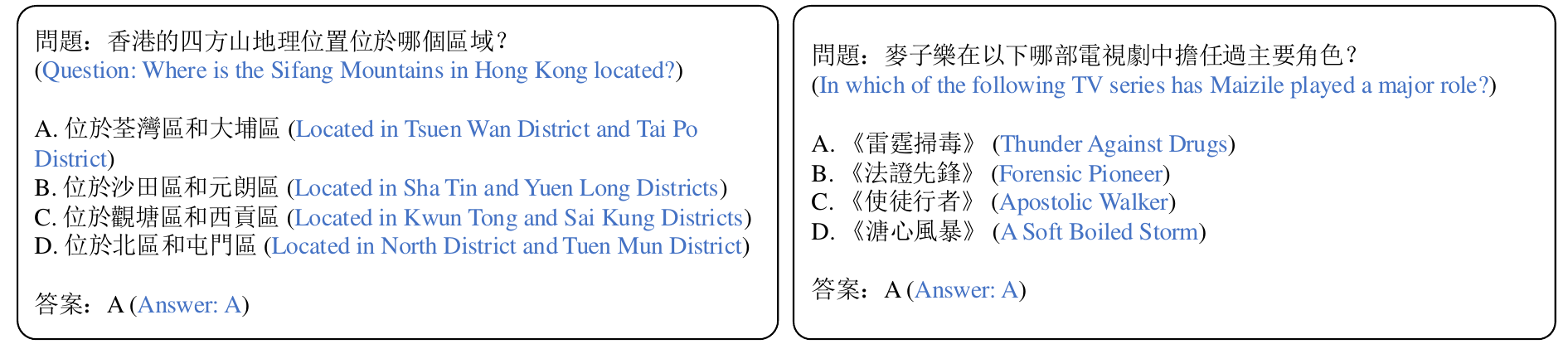}
    \caption{Examples of ``Hong Kong Physical Geography'' and ``Hong Kong Celebrity'' in Hong Kong Specific Data. The \textcolor{blue}{blue} text represents the English translation of the corresponding Chinese.}
    \label{fig:data_example_hk_phygeo_entert}
\end{figure}

\begin{figure}
    \centering
    \includegraphics[width=1\linewidth]{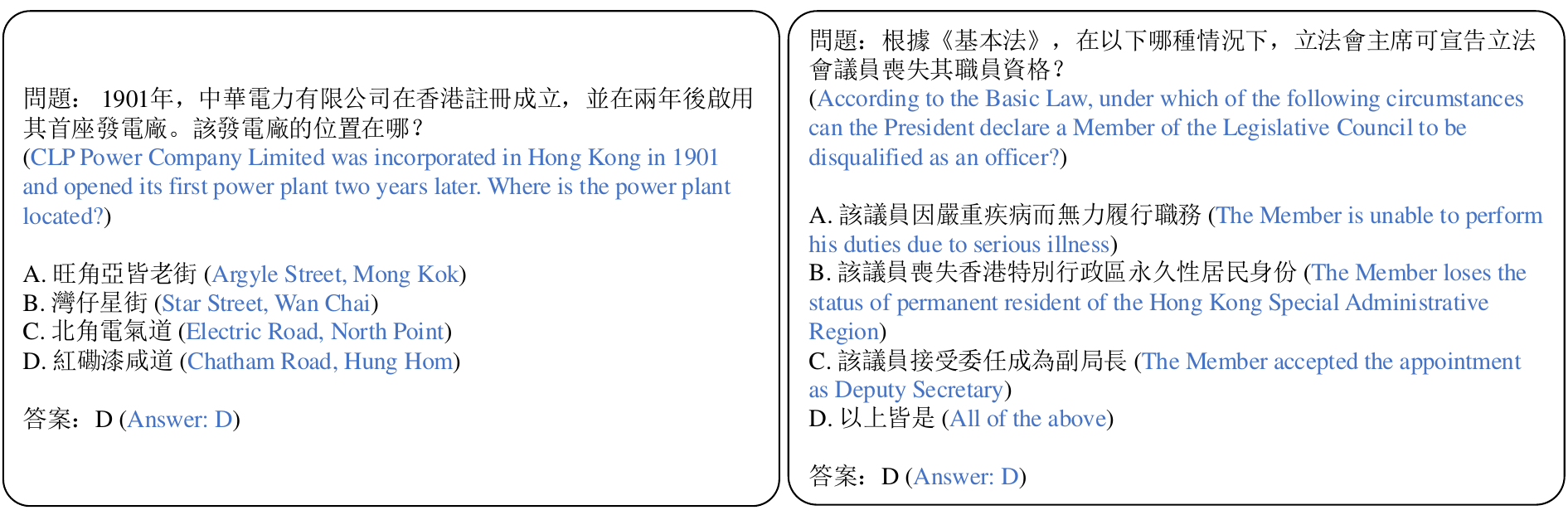}
    \caption{Examples of ``Hong Kong History'' and ``Hong Kong Law'' in Hong Kong Specific Data. The \textcolor{blue}{blue} text represents the English translation of the corresponding Chinese.}
    \label{fig:data_example_hk_hist_law}
\end{figure}

\begin{figure}
    \centering
    \includegraphics[width=\linewidth]{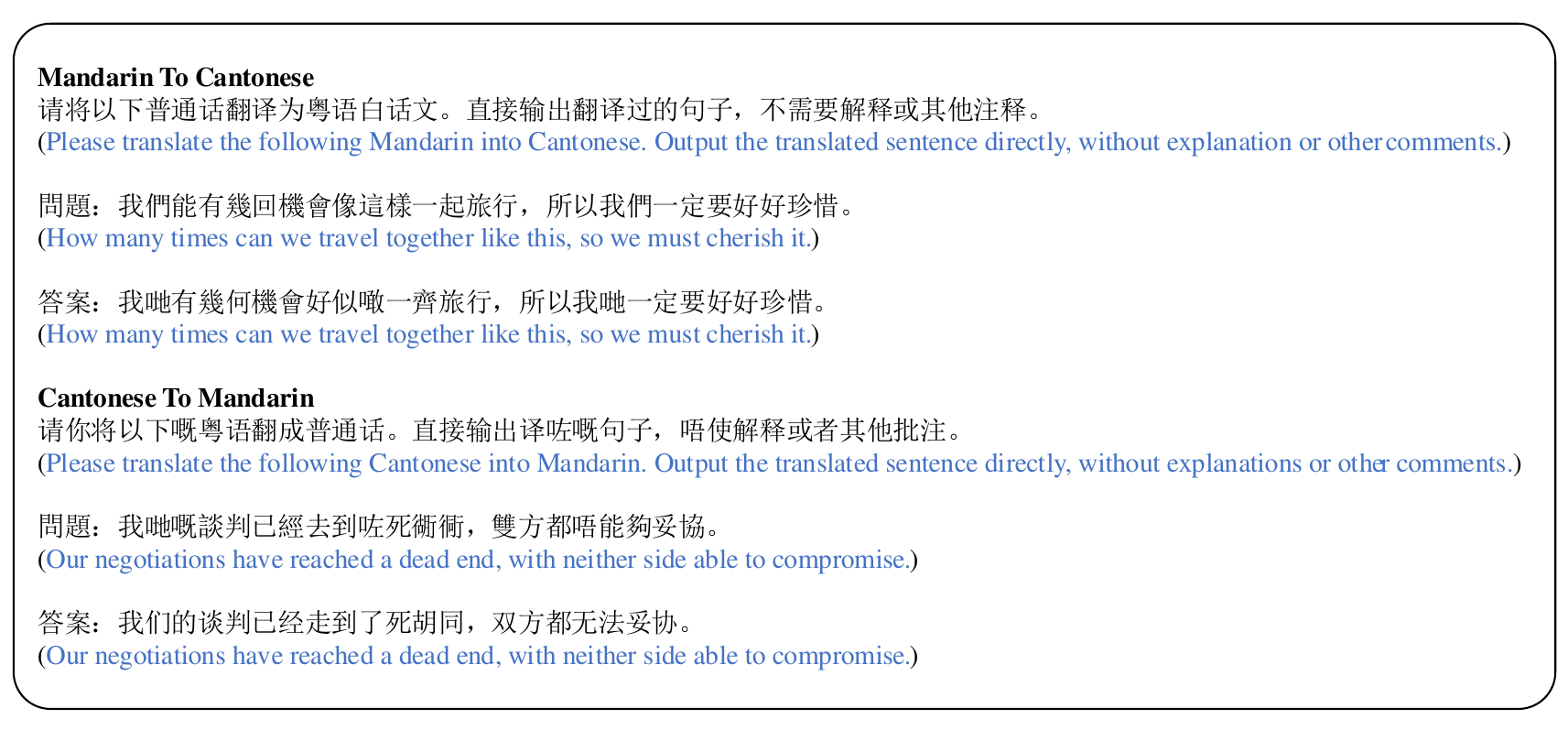}
    \caption{Examples of translation tasks. The \textcolor{blue}{blue} text represents the English translation of the corresponding Chinese.}
    \label{fig:data_example_translate}
\end{figure}

\begin{figure}
    \centering
    \includegraphics[width=\linewidth]{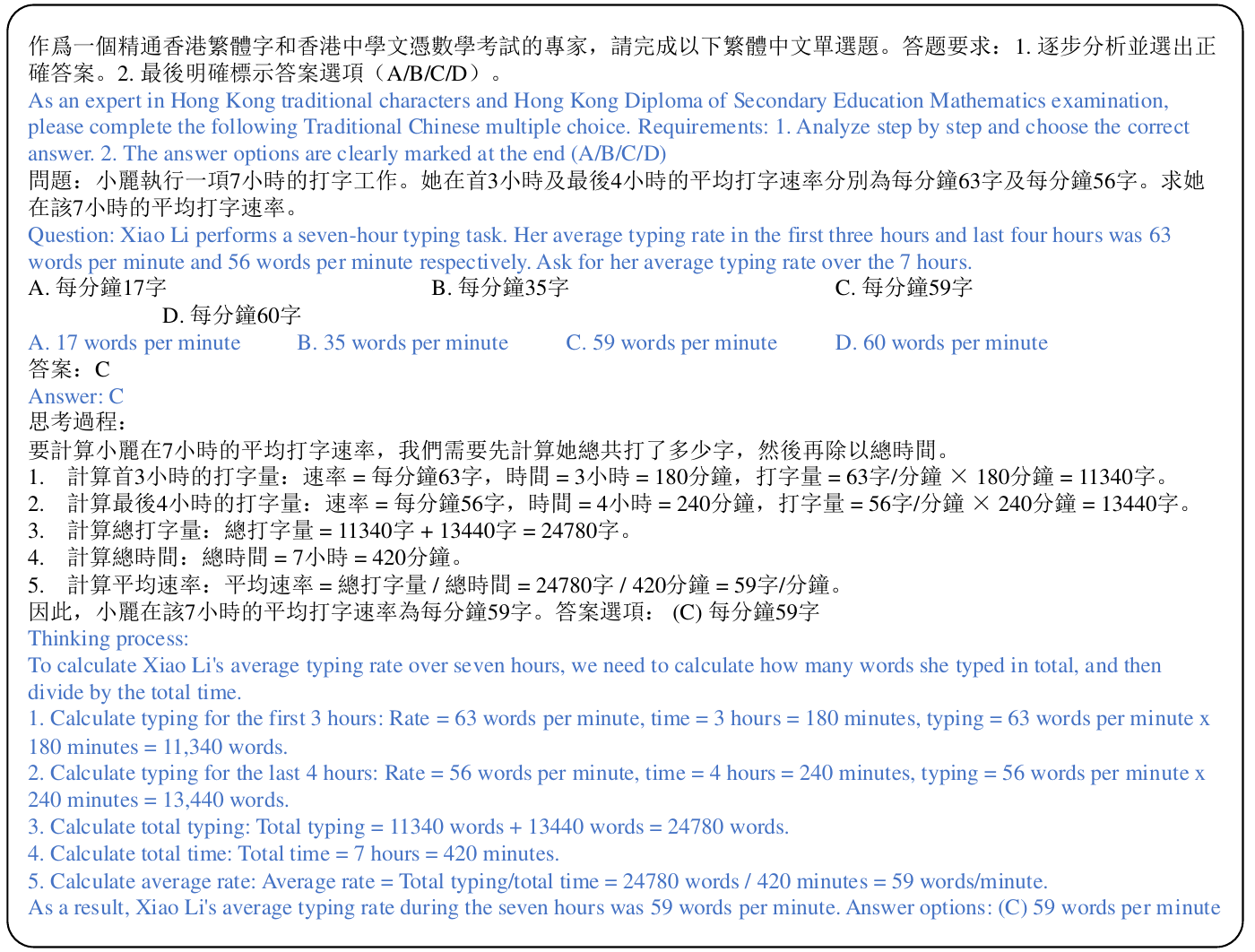}
    \caption{Example of Answering with CoT prompting. The \textcolor{blue}{blue} text represents the English translation of the corresponding Chinese.}
    \label{fig:data_example_cot}
\end{figure}

\begin{figure}
    \centering
    \includegraphics[width=1\linewidth]{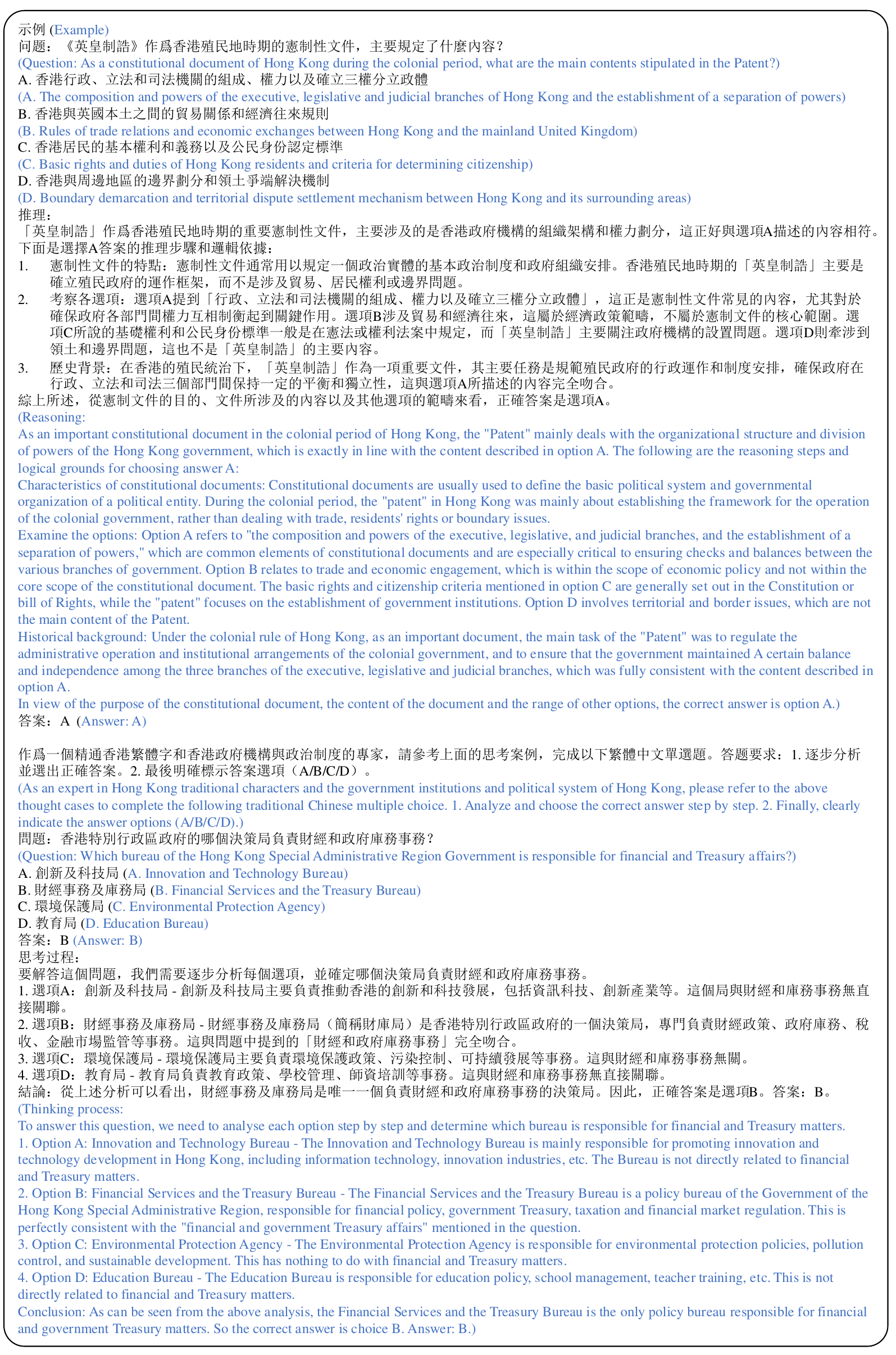}
    \caption{Example of a one-shot CoT prompt. The \textcolor{blue}{blue} text represents the English translation of the corresponding Chinese.}
    \label{fig:data_example_1cot}
\end{figure}

Figures~\ref{fig:data_example_stem_socsci} and Figure~\ref{fig:data_example_human_other} showcase data samples from the four main categories: STEM, Social Sciences, Humanities, and Other. Additionally, Figures~\ref{fig:data_example_hk_phygeo_entert} and~\ref{fig:data_example_hk_hist_law} showcase examples from four Hong Kong-specific topics: physical geography, celebrities, history, and law. Furthermore, Figure~\ref{fig:data_example_translate} provides detailed samples of translation tasks. In CoT prompting, we illustrate complete zero-shot and one-shot CoT examples in Figures~\ref{fig:data_example_cot} and Figure~\ref{fig:data_example_1cot}.

\section{Compute and Resources Used for Evaluation}
For models with API access, including GPT-4o, Claude 3.7 Sonnet, and DeepSeek-V3, we perform inference on CPUs, completing 0-shot tasks within one day. For other models, we utilize a cluster with 8 NVIDIA H800-80GB GPUs and vLLM for acceleration, finishing 0-shot experiments within one day.

\section{Fair and Ethical Labor}

We hired 24 test-takers with a post-secondary degree or higher to participate in the assessment. To recognize their contributions, we established a fair compensation system, offering an estimated average hourly wage of USD 11.58.

\end{document}